\documentclass[10pt,twocolumn,letterpaper]{article}

\usepackage[final]{cvpr}      
\usepackage{amsmath} 
\usepackage{array}
\usepackage{multirow}
\usepackage{graphicx}
\usepackage{caption}
\definecolor{cvprblue}{rgb}{0.21,0.49,0.74}
\usepackage[pagebackref,breaklinks,colorlinks,allcolors=cvprblue]{hyperref}

\definecolor{zxk}{rgb}{0,0,0}
\newcommand{\zxk}[1]{{\color{zxk} {#1}}}

\definecolor{rebuttal}{rgb}{0,1,0}

\definecolor{hsw}{rgb}{0.6,0.2,0.2}

\definecolor{zxkk}{rgb}{0,0,0}
\newcommand{\zxkk}[1]{{\color{zxkk} {#1}}}
\def\paperID{1172} 
\def\confName{CVPR}
\def\confYear{2026}

\title{Learning Explicit Continuous Motion Representation for Dynamic Gaussian Splatting from Monocular Videos}

\author{
Xuankai Zhang\textsuperscript{1}, Junjin Xiao\textsuperscript{1}, Shangwei Huang\textsuperscript{1}, Wei-shi Zheng\textsuperscript{1,2}, Qing Zhang\textsuperscript{1,2}\thanks{Corresponding author (zhangq93@mail.sysu.edu.cn).}  \\
\textsuperscript{1} School of Computer Science and Engineering, Sun Yat-sen University, China \\
\textsuperscript{2}Key Laboratory of Machine Intelligence and Advanced Computing, Ministry of Education, China \\
}

\begin{document}

\twocolumn[
    \maketitle
    \vspace{-3.0em}
    \begin{center}
    \includegraphics[width=0.99\textwidth]{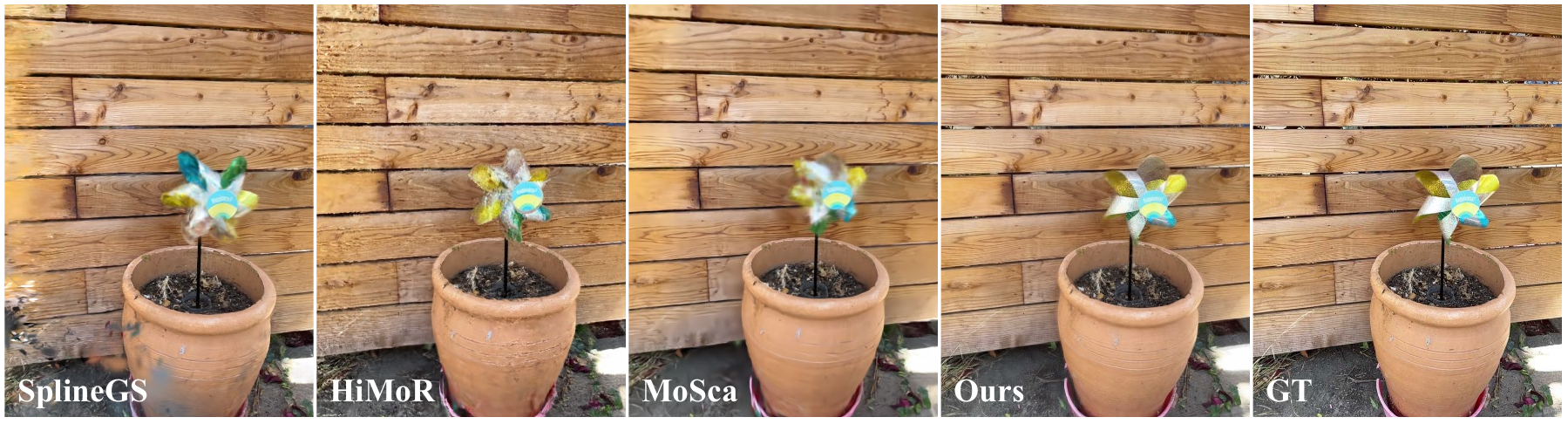}\\
    \vspace{-2mm}
    \captionof{figure}{
        \textbf{Dynamic Gaussian Splatting from monocular videos.} Our method synthesizes high-quality novel views from monocular videos, while the compared methods, e.g., MoSca~\cite{mosca}, HiMoR~\cite{himor}, and SplineGS~\cite{splinegs}, fail to faithfully reconstruct the dynamic windmill. 
    }
    \label{fig:teaser}
\end{center}
    \bigbreak
]

\begingroup
\renewcommand\thefootnote{}
\footnotetext{*Corresponding author (zhangq93@mail.sysu.edu.cn).}
\endgroup

\begin{abstract}
\label{sec:abstract}

We present an approach for high-quality dynamic Gaussian Splatting from monocular videos. To this end, we in this work go one step further beyond previous methods to explicitly model continuous position and orientation deformation of dynamic Gaussians, using an SE(3) B-spline motion bases with a compact set of control points. To improve computational efficiency while enhancing the ability to model complex motions, an adaptive control mechanism is devised to dynamically adjust the number of motion bases and control points. Besides, we develop a soft segment reconstruction strategy to mitigate long-interval motion interference, and employ a multi-view diffusion model to provide multi-view cues for avoiding overfitting to training views. Extensive experiments demonstrate that our method outperforms state-of-the-art methods in novel view synthesis. Our code is available at \href{https://github.com/hhhddddddd/se3bsplinegs}{\textcolor{cyan}{https://github.com/hhhddddddd/se3bsplinegs}}.


\end{abstract}    
\section{Introduction}
\label{sec:intro}

Novel view synthesis (NVS) is a fundamental problem in computer vision with various applications in virtual reality, augmented reality, and film production. Although significant progress has been made to synthesize novel views from multi-view dynamic videos, it remains a challenge to reconstruct high-quality novel views from monocular videos lack of multi-view cues.


Building upon 3D Gaussian Splatting (3DGS) \cite{3dgs} and Neural Radiance Field (NeRF) \cite{nerf}, numerous methods have been proposed to reconstruct dynamic scenes from monocular videos. Some of them \cite{d3dgs,som,himor,hexplane} implicitly construct dynamic Gaussian deformation trajectories by learning a Gaussian transformation from canonical space to observation space. On the other hand, SplineGS \cite{splinegs} proposes to explicitly models the continuous Gaussian position deformation trajectories via a cubic hermite spline, which is subsequently extended to improve the computational efficiency \cite{nodegs}. Although these methods demonstrate promising results, they are susceptible to non-continuous Gaussian orientation deformation and tend to incur artifacts, especially in regions with complex motions. 

To address the limitation of previous methods, we in this work present an approach that allows high-quality dynamic Gaussian Splatting from monocular videos, by explicitly modeling continuous Gaussian position and orientation deformation trajectories. To do so, we employ an explicit continuous SE(3) B-spline motion bases for controlling dynamic Gaussian trajectories using the SE(3) cumulative B-spline with a small number of control points. To improve computational efficiency while enhancing robustness to complex motions, we devise an adaptive control mechanism to dynamically adjust the number of motion bases and control points. Moreover, we develop a soft segment reconstruction strategy to reduce the interference of dynamic Gaussians led by long-interval motion deformation during scene reconstruction. Finally, we adopt multi-view cues from a multi-view diffusion model to avoid overfitting to training views by imposing a SDS loss. 

In summary, our main contributions are:


\begin{itemize}[leftmargin=2em]
\setlength\itemsep{0.5em} 
  \item We present a dynamic Gaussian Splatting framework that explicitly models continuous position and orientation deformation of dynamic Gaussians using adaptive SE(3) B-spline motion bases.
  
  
  \item We introduce a soft segment reconstruction strategy to reduce artifacts from dynamic Gaussians under long-term deformation, and propose to employ multi-view cues from multi-view diffusion model to enable better generalization to novel views. 
  
  \item Experiments on benchmark monocular video datasets demonstrate the superiority of our method over state-of-the-art novel view synthesis methods.
\end{itemize}

\section{Related Works}
\label{sec:relatedwork}

\noindent \textbf{Dynamic NeRF.} Neural Radiance Field (NeRF) has shown great success in novel view synthesis for static scenes. Recent works extend NeRF to handle dynamic scenes, which can be broadly divided into two categories. One line of works uses time-varying neural radiance fields to model dynamic scenes \cite{hexplane, k-planes, tensor4d, hyperreel}. For example, K-Planes \cite{k-planes} introduces a spatiotemporal radiance field interpolating the feature vectors indexed by time. Another line of works model dynamic scene using a canonical space NeRF and a deformation field \cite{nerfies, hypernerf, d-nerf, nonerf, volumetric, humannerf, neuman, banmo}. Among them, HyperNeRF \cite{hypernerf} models the scene dynamics as a deformation field mapping to a canonical space. However, existing dynamic NeRF methods still suffer from low rendering efficiency due to dense ray sampling and expensive volume rendering, and struggle to produce high-quality novel views from monocular videos.

\vspace{0.5em}
\noindent \textbf{Dynamic 3DGS.} There are various methods focusing on achieving dynamic Gaussian Splatting by introducing learnable deformations to Gaussian \cite{4dgsroter,4dgsproject,dreamscene4d,modgs,rogsplat,d3dgs,spacetimegs,dynamic-gs}. One line of works define motion as a time-conditioned deformation network that warps Gaussians from canonical space to observation space \cite{4dgscnn,d3dgs,4dscaffold,motiongs,uags,sc-gs,gaufre}. Among them, D3DGS \cite{d3dgs} models motion with an MLP that predicts deformation variables between canonical space and observation space. 4DGS \cite{4dgscnn} replaces MLP with k-plane \cite{k-planes} to extract spatio-temporal information that is beneficial to dynamic reconstruction. Another line of works model motion as trajectories of 3D Gaussians \cite{trackgs,spacetimegs,dydeblur,splinegs,ane3d,e3dgs,som,3dgstream,himor}. For instance, Shape-of-Motion \cite{som} models motion as a linear combination of $\mathbb{SE}(3)$ motion bases, while MoSca \cite{mosca} uses 4D motion scaffolds built upon 2D foundational models as the motion representation. However, as these methods do not explicitly model continuous Gaussian position and orientation deformation trajectories in a unified framework, they are less effective in dealing with monocular videos with complex motions. 


\section{Method}
\label{sec:method}


 \begin{figure*}
     \centering
     \includegraphics[width=1.0\linewidth]{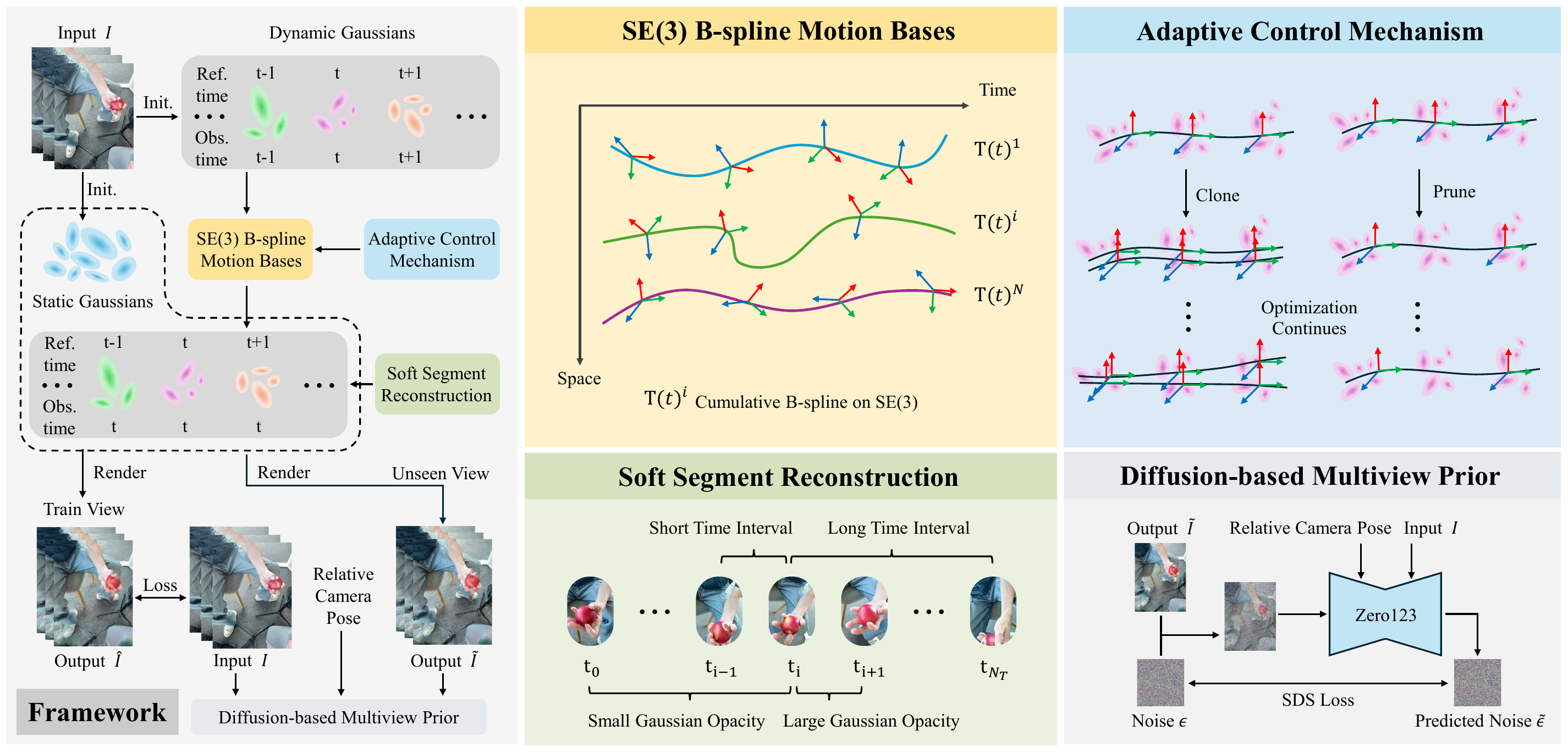}\\
      \vspace{-2mm}
     \caption{\textbf{Overview of our method.} We first initialize static Gaussians via depth reprojection and dynamic Gaussians from tracking points, by modeling their transformations with learnable SE(3) B-spline Motion Bases. We then adjust the number of motion bases and control points based on an adaptive control mechanism. \zxk{Next, we employ a soft segment reconstruction strategy to fuse dynamic Gaussians at different reference timestamps to the observation timestamp}, and further supplement the monocular video with scene-level multi-view cues derived from a multi-view diffusion model.}
     \label{fig:overview}
     \vspace{-5pt}
 \end{figure*}



\zxk{Our method aims to achieve dynamic Gaussian splatting that enables high-fidelity yet sharp novel view synthesis from monocular videos. Figure \ref{fig:overview} presents an overview of our method. Below we describe our method in detail.}

\subsection{SE(3) B-spline Motion Bases}
\label{sec:SE(3)}
\noindent \textbf{Representation of motion base.}
\zxk{Current methods for reconstructing dynamic scenes from monocular video do not explicitly model continuous dynamic Gaussian deformation trajectories. Some existing methods \cite{som,himor} model deformation by learning the corresponding affine transformations from the canonical space to the observation space at each timestamp. These cannot guarantee that the dynamic Gaussian maintains a continuous trajectory within the observation space at each timestamp. Although other methods \cite{splinegs,nodegs} model the continuous position trajectory of the dynamic Gaussian using Cubic Hermite Spline function \cite{hermite}, they cannot guarantee that the orientation changes of the dynamic Gaussian are continuous. This results in non-smooth pose variations of the dynamic Gaussian, leading to significant artifacts in the rendered images.}

\zxk{To ensure that the dynamic Gaussian maintains continuous positional and orientational deformations, we represent the motion bases of dynamic objects using mathematically continuous SE(3) Cumulative B-spline \cite{bspline}. Then, we use SE(3) B-spline motion bases to control the dynamic Gaussians so that they exhibit continuous deformation in both position and orientation. Specifically, the SE(3) B-spline motion bases are constructed by learnable control points. To initialize control points, we follow the data preprocessing procedure of MoSca \cite{mosca} to obtain the pose state $Q = [R, t]$ of 3D tracklets, where $t$ represents their spatial position and $R$ their orientation in 3D space. Then, we utilize the pose state $Q$ as the control points. Similar to SE(3) Cumulative B-spline\cite{bspline}, we compute the relative pose transformation between adjacent 3D tracklets as follows:}
\begin{equation}
  \Delta Q = Q_i^{-1}Q_{i+1},
  \label{eq:relativeQ}
\end{equation}
\zxk{where $Q_i$ and $Q_{i+1}$ denote the pose state of the $i$-th and $(i+1)$-th 3D tracklets, respectively. Then, we transform the relative pose transformation $\Delta Q$ into the Lie algebraic space \cite{li} using the Lie algebraic logarithmic transformation as follows:}
\begin{equation}
  \xi = log(\Delta Q),
  \label{eq:logrelativeQ}
\end{equation}
\zxk{where $\xi$ denotes the Lie algebra space relative pose transformation, which represents the speed of the pose transformation between adjacent 3D tracklets. Based on SE(3) Cumulative B-spline \cite{bspline}, we define the representation of SE(3) B-spline motion bases as follows:}
\begin{equation}
  T(t) = \big( \displaystyle\prod_{i=0}^{N_c-1}exp(\Omega_i(t) \xi_i)  \big) T_0,
  \label{eq:se3motion}
\end{equation}
\zxk{where $\Omega_i(t)$ represents the B-spline basis function, $N_c$ denotes the number of control points, and $T_0$ represents the pose state of the 3D tracklets in the first frame. In the supplementary material, we introduce the initialization of SE(3) B-spline motion bases.}

\vspace{0.5em}
\noindent \textbf{Adaptive control of motion bases.}
\zxk{Excessive control points not only increase the computational complexity of deformation but also risk causing overfitting of the SE(3) B-spline motion bases, leading to degraded rendering efficiency and quality. Moreover, since different regions of the scene exhibit diverse motion complexity, the number of motion bases should be adaptively densified according to local motion complexity. To address this, we introduce an adaptive control mechanism for SE(3) B-spline motion bases, which prunes redundant control points and increases the density of motion bases adaptively without compromising expressive power.}

\zxk{For pruning control points, we perform a pruning operation every $N_{prune}$ iterations. Each pruning operation attempts to reduce the control points by one. Specifically, we first select the optimal pruning control point in each motion base as follows:}
\begin{equation}
    \min_{{Q}} \textstyle \sum^{N_T}_{t=0} \big\|T(t)^{Q} - T(t) \big\|^2_2,
  \label{eq:prune}
\end{equation}
\zxk{where $N_T$ represents the number of monocular video frames, $T(t)^{Q}$ represents the SE(3) B-spline motion base after removing the control point $Q$, and $T(t)$ represents the original SE(3) B-spline motion base. Then, we remove the optimal pruning point $\hat Q$ if the error $E$ between $T(t)^{Q}$ and $T(t)$ is smaller than a threshold $\epsilon_{prune}=5.0$, $E$ is calculated as follows:}
\begin{equation}
     E= \textstyle \sum^{N_T}_{t=0} \big\|T(t)^{\hat Q} - T(t) \big\|^2_2.
  \label{eq:pruneror}
\end{equation}


\begin{figure*}[!t]
    \centering
    \setlength{\tabcolsep}{1.0pt}
    \scriptsize
    \begin{tabular}{cccccccc}
        \begin{minipage}[b]{0.12000\linewidth}
        \includegraphics[width=1\linewidth]
        {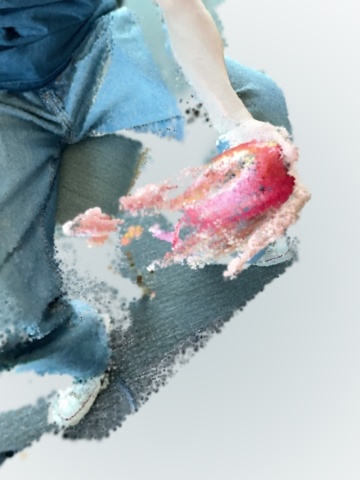}
        \end{minipage}
        &  
        \begin{minipage}[b]{0.12000\linewidth}
        \includegraphics[width=1\linewidth]{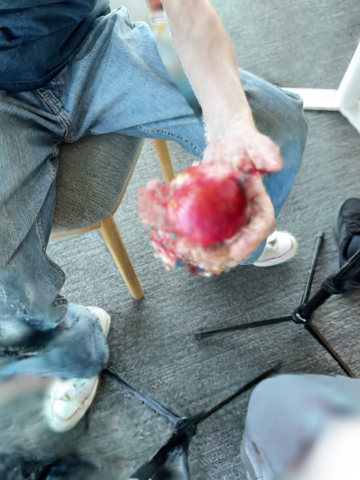}
        \end{minipage}
        &  
        \begin{minipage}[b]{0.12000\linewidth}
        \includegraphics[width=1\linewidth]{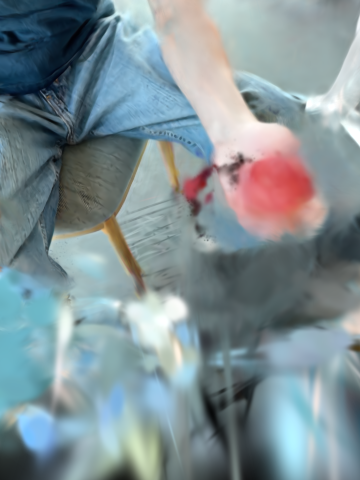} 
        \end{minipage}
        &  
        \begin{minipage}[b]{0.12000\linewidth}
        \includegraphics[width=1\linewidth]{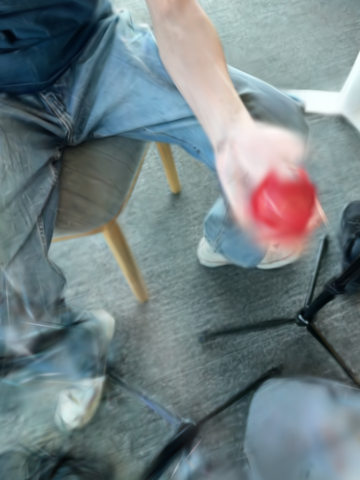} 
        \end{minipage}
        &  
        \begin{minipage}[b]{0.12000\linewidth}
        \includegraphics[width=1\linewidth]{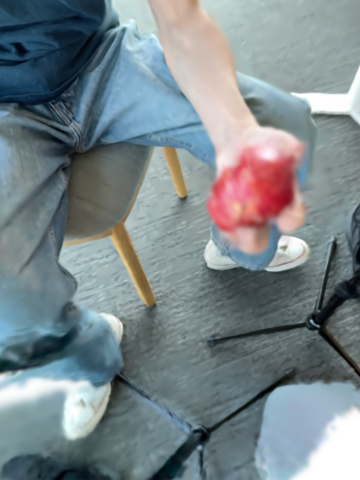} 
        \end{minipage}
        &  
        \begin{minipage}[b]{0.12000\linewidth}
        \includegraphics[width=1\linewidth]{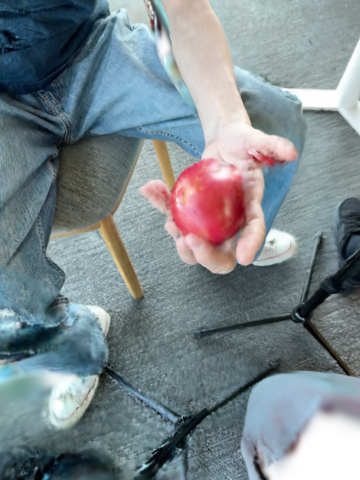} 
        \end{minipage}
        &  
        \begin{minipage}[b]{0.12000\linewidth}
        \includegraphics[width=1\linewidth]{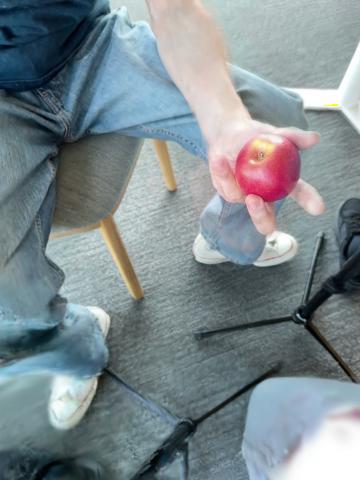} 
        \end{minipage}        
        &  
        \begin{minipage}[b]{0.12000\linewidth}
        \includegraphics[width=1\linewidth]{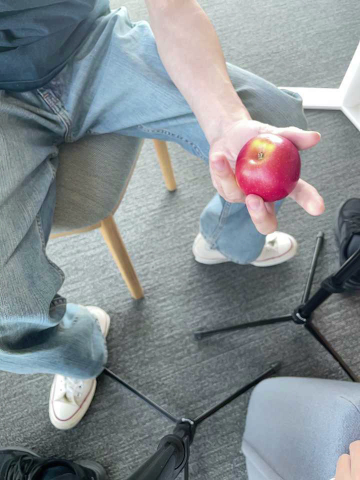}
        \end{minipage}
        \\
        \begin{minipage}[b]{0.12000\linewidth}
        \includegraphics[width=1\linewidth]{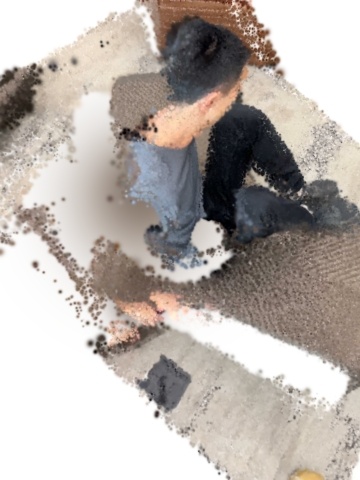}
        \end{minipage}
        &  
        \begin{minipage}[b]{0.12000\linewidth}
        \includegraphics[width=1\linewidth]{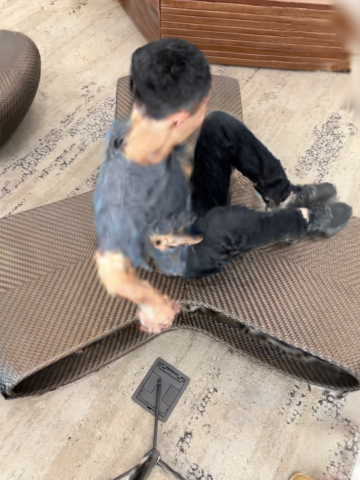} 
        \end{minipage}
        &  
        \begin{minipage}[b]{0.12000\linewidth}
        \includegraphics[width=1\linewidth]{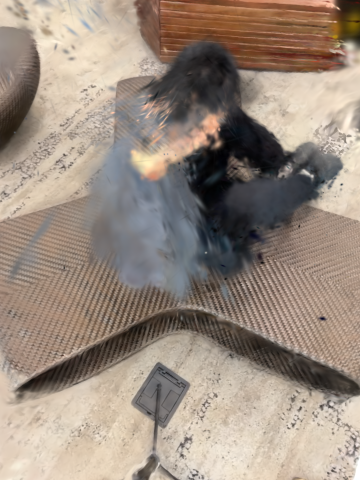} 
        \end{minipage}
        &  
        \begin{minipage}[b]{0.12000\linewidth}
        \includegraphics[width=1\linewidth]{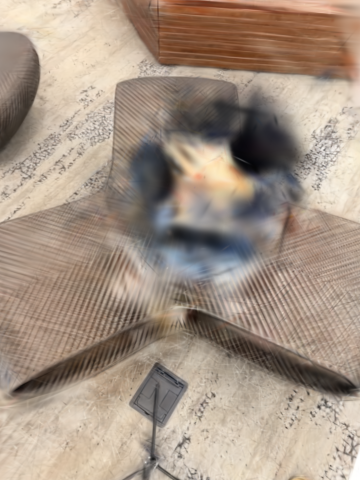} 
        \end{minipage}
        &  
        \begin{minipage}[b]{0.12000\linewidth}
        \includegraphics[width=1\linewidth]{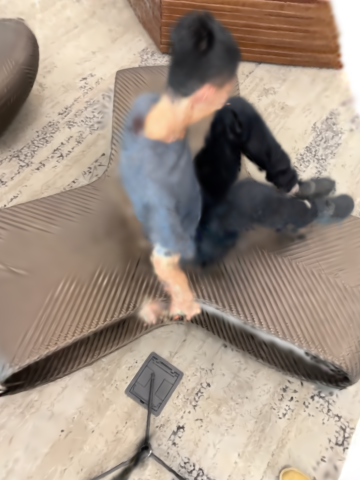} 
        \end{minipage}
        &  
        \begin{minipage}[b]{0.12000\linewidth}
        \includegraphics[width=1\linewidth]{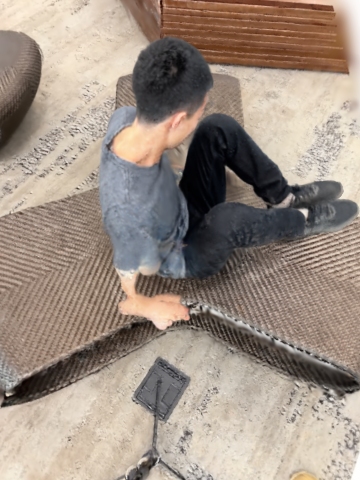} 
        \end{minipage}
        &  
        \begin{minipage}[b]{0.12000\linewidth}
        \includegraphics[width=1\linewidth]{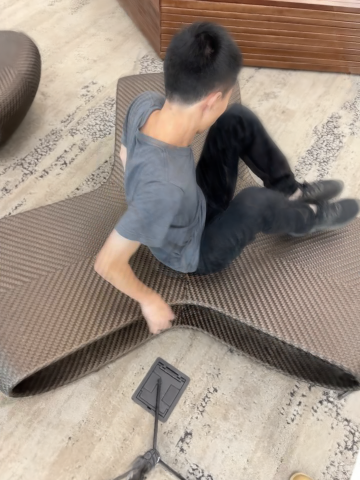}
        \end{minipage}
        &  
        \begin{minipage}[b]{0.12000\linewidth}
        \includegraphics[width=1\linewidth]{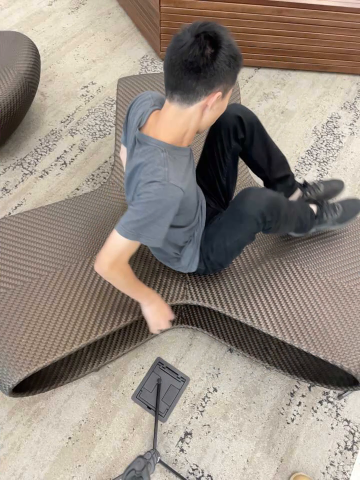}
        \end{minipage}
        \\ 
        \begin{minipage}[b]{0.12000\linewidth}
        \includegraphics[width=1\linewidth]{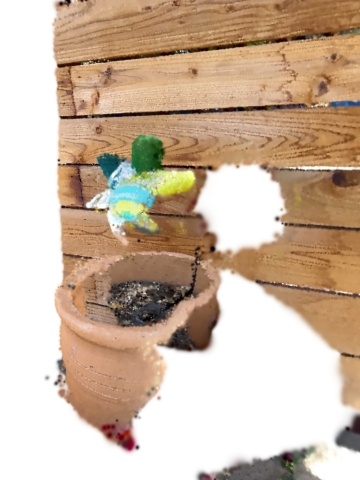}
        \end{minipage}
        &  
        \begin{minipage}[b]{0.12000\linewidth}
        \includegraphics[width=1\linewidth]{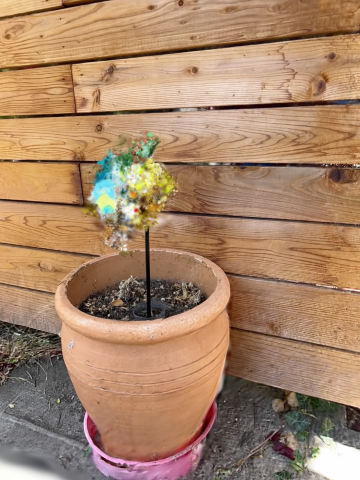} 
        \end{minipage}
        &  
        \begin{minipage}[b]{0.12000\linewidth}
        \includegraphics[width=1\linewidth]{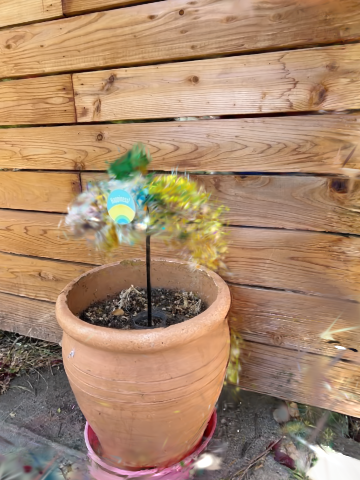} 
        \end{minipage}
        &  
        \begin{minipage}[b]{0.12000\linewidth}
        \includegraphics[width=1\linewidth]{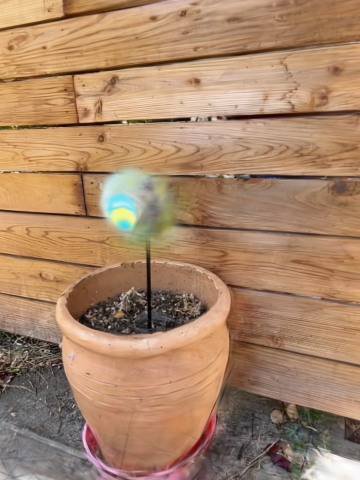} 
        \end{minipage}
        &  
        \begin{minipage}[b]{0.12000\linewidth}
        \includegraphics[width=1\linewidth]{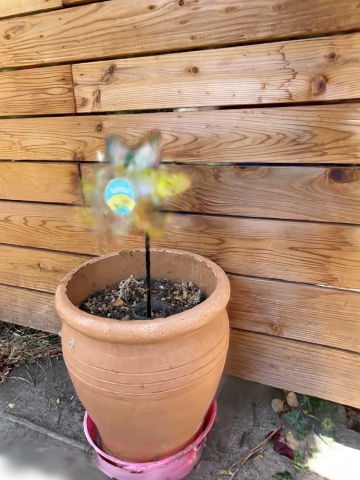} 
        \end{minipage}
        &  
        \begin{minipage}[b]{0.12000\linewidth}
        \includegraphics[width=1\linewidth]{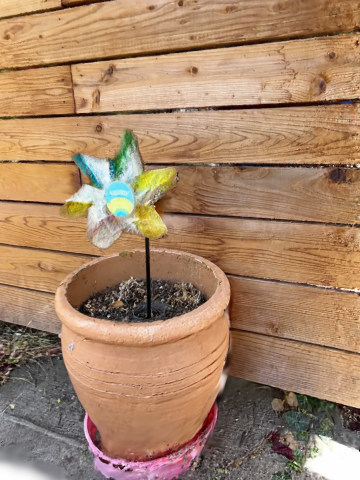} 
        \end{minipage}
        &  
        \begin{minipage}[b]{0.12000\linewidth}
        \includegraphics[width=1\linewidth]{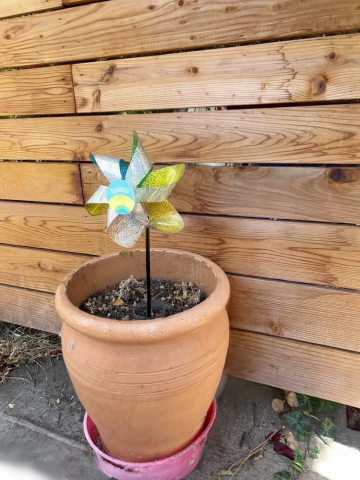}
        \end{minipage}
        &  
        \begin{minipage}[b]{0.12000\linewidth}
        \includegraphics[width=1\linewidth]{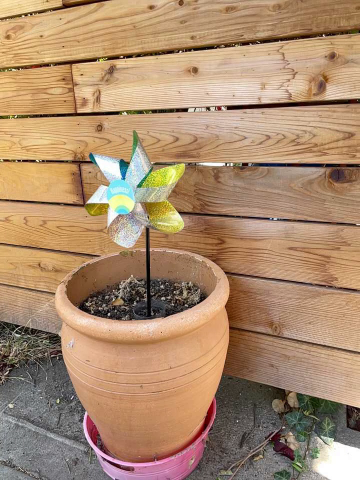}
        \end{minipage}
        \\ 
        \begin{minipage}[b]{0.12000\linewidth}
        \includegraphics[width=1\linewidth]{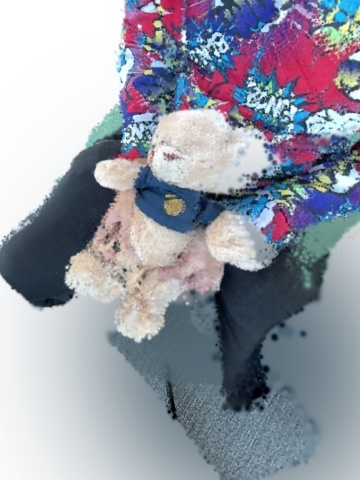}
        \end{minipage}
        &  
        \begin{minipage}[b]{0.12000\linewidth}
        \includegraphics[width=1\linewidth]{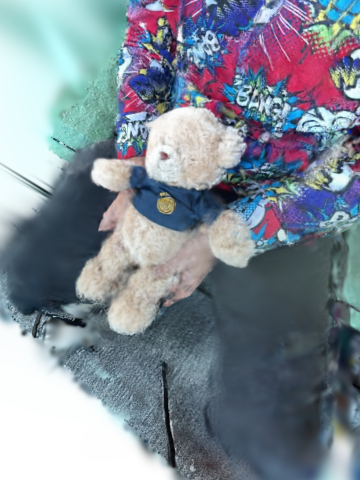} 
        \end{minipage}
        &  
        \begin{minipage}[b]{0.12000\linewidth}
        \includegraphics[width=1\linewidth]{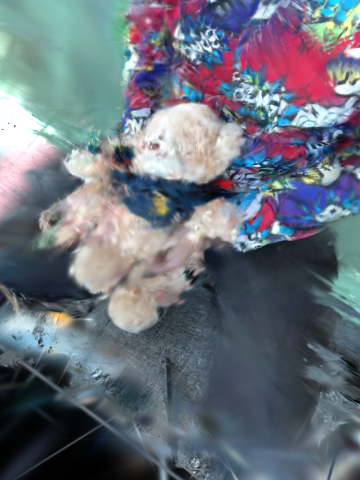} 
        \end{minipage}
        &  
        \begin{minipage}[b]{0.12000\linewidth}
        \includegraphics[width=1\linewidth]{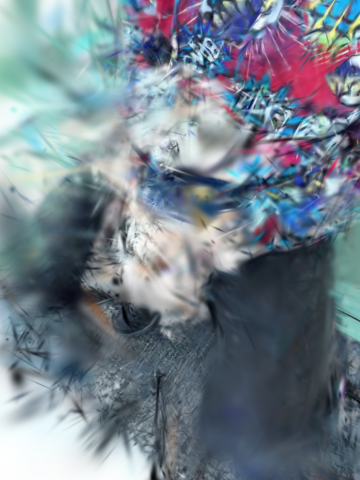} 
        \end{minipage}
        &  
        \begin{minipage}[b]{0.12000\linewidth}
        \includegraphics[width=1\linewidth]{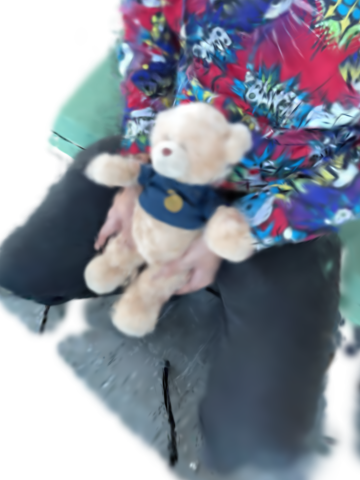} 
        \end{minipage}
        &  
        \begin{minipage}[b]{0.12000\linewidth}
        \includegraphics[width=1\linewidth]{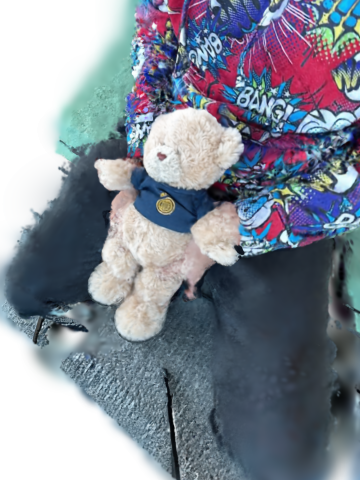} 
        \end{minipage}
        &  
        \begin{minipage}[b]{0.12000\linewidth}
        \includegraphics[width=1\linewidth]{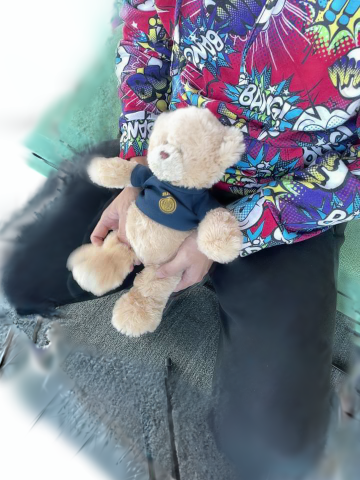}
        \end{minipage}
        &  
        \begin{minipage}[b]{0.12000\linewidth}
        \includegraphics[width=1\linewidth]{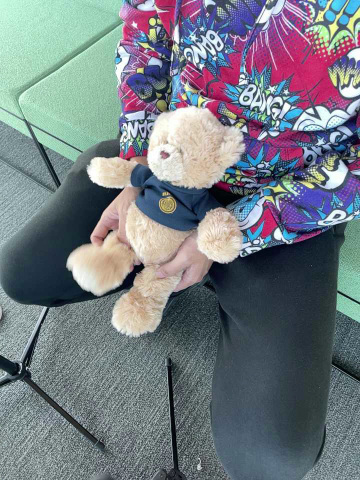}
        \end{minipage}
        \\ 
        \makebox[0.12000\linewidth][c]{\footnotesize MarbleGS~\cite{marblegs}} & 
        \makebox[0.12000\linewidth][c]{\footnotesize SoM~\cite{som}} & 
        \makebox[0.12000\linewidth][c]{\footnotesize SplineGS~\cite{splinegs}} & 
        \makebox[0.12000\linewidth][c]{\footnotesize MoDec-GS~\cite{modec-gs}} & 
        \makebox[0.12000\linewidth][c]{\footnotesize MoSca~\cite{mosca}} & 
        \makebox[0.12000\linewidth][c]{\footnotesize HiMoR~\cite{himor}} & 
        \makebox[0.12000\linewidth][c]{\footnotesize Ours} & 
        \makebox[0.12000\linewidth][c]{\footnotesize GT} \\
    \end{tabular}
    \vspace{-2mm}
    \caption{\textbf{Visual comparison of novel view synthesis on the iPhone dataset \cite{iphone}.}}
    \label{fig:comiphone}
\end{figure*}

\zxk{Every $N_{densify}$ iterations, we perform motion bases densification in complex motion regions where the modeling capacity of motion bases is insufficient. First, we determine the complex motion regions by using the rendering errors at the training views and the corresponding dynamic region masks. Specifically, for the $i$-th training view $v_i$, we obtain the rendering error $m_{error}^i$ as follows:}
\begin{equation}
     m_{error}^i= \begin{cases}
                \textbf{True}, ~ \text{if $|\hat I^i -I^i| > \epsilon_{error}$}\\
                    \textbf{False}, ~ \text{otherwise}
                \end{cases}, \\ 
  \label{eq:densifyerrormask}
\end{equation}
\zxk{where $\hat I^i$ and $I^i$ represent the rendered image and the ground-truth image at training view $v_i$, respectively, and $\epsilon_{error}=0.5$. We obtain the complex motion regions $m^i$ from the current view $v_i$ as follows:}
\begin{equation}
     m^i= m_{error}^i \cap m_{d}^i,
  \label{eq:densifymask}
\end{equation}
\zxk{where $m_d^i$ represents the dynamic region mask at the current view $v_i$, which is obtained using the dynamic region mask calculation method adopted by MoSca \cite{mosca}. According to Eq.~(\ref{eq:densifyerrormask}) and Eq.~(\ref{eq:densifymask}), we obtain the mask sequence $M=[m^0,m^1,...m^{N_T-1}]$ of the complex motion regions at all time stamps. We use $M$ to select the SE(3) B-spline motion bases that need to be densified. Specifically, for the $j$-th SE(3) B-spline motion base $T(t)^j$, we project the 3D position of $T(t)^j$ at all timestamps to the image plane under the corresponding view as follows:}
\begin{equation}
     p= KP(t)T(t)^j,
  \label{eq:world2pixel}
\end{equation}
\zxk{where $p$ denotes the 2D projection of the 3D position of $T(t)^j$ onto the image plane, $K$ represents the camera intrinsics, and $P(t)$ represents the camera extrinsics. If the number of $p$ instances that fall within the mask sequence $M$ exceeds 50\%, we perform a densification operation on $T(t)^j$. Specifically, we simply copy the control point parameters of the SE(3) B-spline motion base $T(t)^j$ and add random perturbations to $T_0^j$.}

\begin{figure*}[!t]
    \centering
    \setlength{\tabcolsep}{1.0pt}
    \scriptsize
    \begin{tabular}{cccccccc}
        \begin{minipage}[b]{0.12000\linewidth}
        \includegraphics[width=1\linewidth]
        {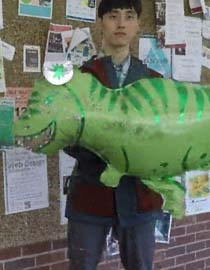}
        \end{minipage}
        &  
        \begin{minipage}[b]{0.12000\linewidth}
        \includegraphics[width=1\linewidth]{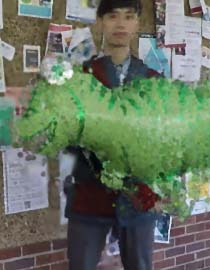}
        \end{minipage}
        &  
        \begin{minipage}[b]{0.12000\linewidth}
        \includegraphics[width=1\linewidth]{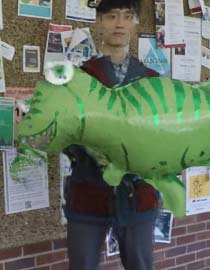} 
        \end{minipage}
        &  
        \begin{minipage}[b]{0.12000\linewidth}
        \includegraphics[width=1\linewidth]{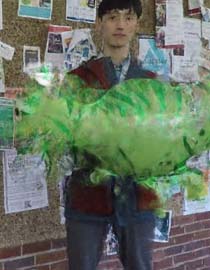} 
        \end{minipage}
        &  
        \begin{minipage}[b]{0.12000\linewidth}
        \includegraphics[width=1\linewidth]{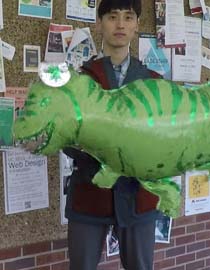} 
        \end{minipage}
        &  
        \begin{minipage}[b]{0.12000\linewidth}
        \includegraphics[width=1\linewidth]{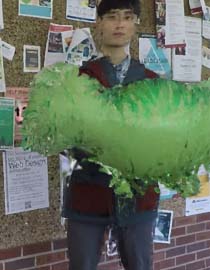} 
        \end{minipage}
        &  
        \begin{minipage}[b]{0.12000\linewidth}
        \includegraphics[width=1\linewidth]{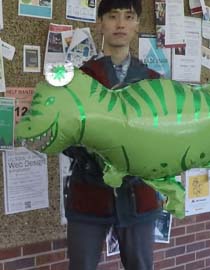} 
        \end{minipage}        
        &  
        \begin{minipage}[b]{0.12000\linewidth}
        \includegraphics[width=1\linewidth]{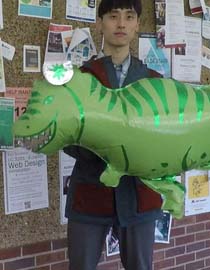}
        \end{minipage}
        \\
        \begin{minipage}[b]{0.12000\linewidth}
        \includegraphics[width=1\linewidth]{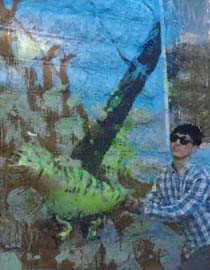}
        \end{minipage}
        &  
        \begin{minipage}[b]{0.12000\linewidth}
        \includegraphics[width=1\linewidth]{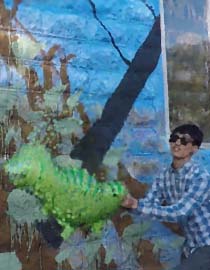} 
        \end{minipage}
        &  
        \begin{minipage}[b]{0.12000\linewidth}
        \includegraphics[width=1\linewidth]{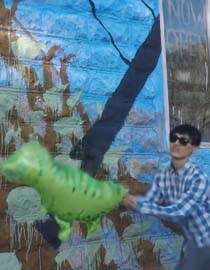} 
        \end{minipage}
        &  
        \begin{minipage}[b]{0.12000\linewidth}
        \includegraphics[width=1\linewidth]{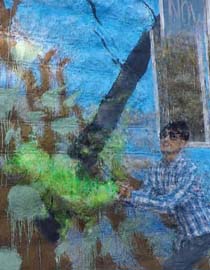} 
        \end{minipage}
        &  
        \begin{minipage}[b]{0.12000\linewidth}
        \includegraphics[width=1\linewidth]{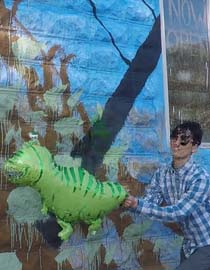} 
        \end{minipage}
        &  
        \begin{minipage}[b]{0.12000\linewidth}
        \includegraphics[width=1\linewidth]{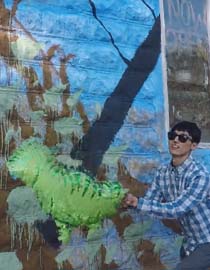} 
        \end{minipage}
        &  
        \begin{minipage}[b]{0.12000\linewidth}
        \includegraphics[width=1\linewidth]{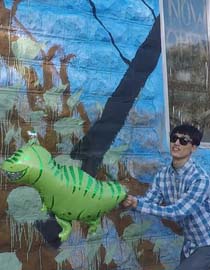}
        \end{minipage}
        &  
        \begin{minipage}[b]{0.12000\linewidth}
        \includegraphics[width=1\linewidth]{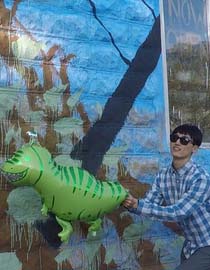}
        \end{minipage}
        \\ 
        \begin{minipage}[b]{0.12000\linewidth}
        \includegraphics[width=1\linewidth]{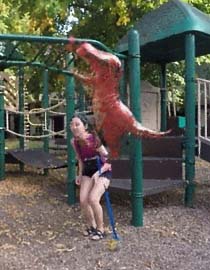}
        \end{minipage}
        &  
        \begin{minipage}[b]{0.12000\linewidth}
        \includegraphics[width=1\linewidth]{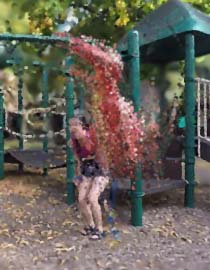} 
        \end{minipage}
        &  
        \begin{minipage}[b]{0.12000\linewidth}
        \includegraphics[width=1\linewidth]{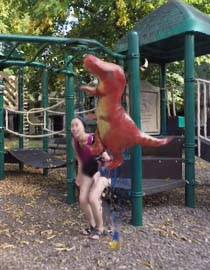} 
        \end{minipage}
        &  
        \begin{minipage}[b]{0.12000\linewidth}
        \includegraphics[width=1\linewidth]{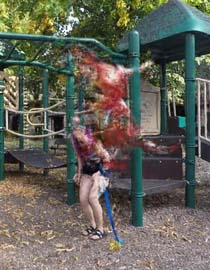} 
        \end{minipage}
        &  
        \begin{minipage}[b]{0.12000\linewidth}
        \includegraphics[width=1\linewidth]{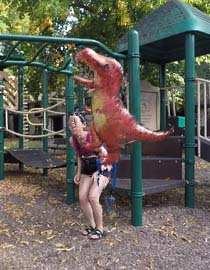} 
        \end{minipage}
        &  
        \begin{minipage}[b]{0.12000\linewidth}
        \includegraphics[width=1\linewidth]{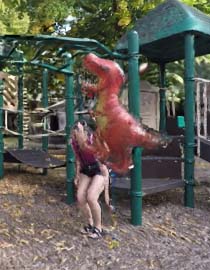} 
        \end{minipage}
        &  
        \begin{minipage}[b]{0.12000\linewidth}
        \includegraphics[width=1\linewidth]{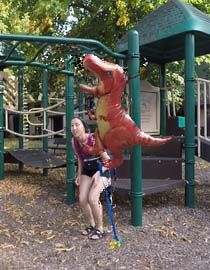}
        \end{minipage}
        &  
        \begin{minipage}[b]{0.12000\linewidth}
        \includegraphics[width=1\linewidth]{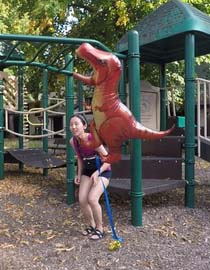}
        \end{minipage}
        \\ 
        \begin{minipage}[b]{0.12000\linewidth}
        \includegraphics[width=1\linewidth]{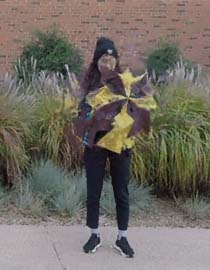}
        \end{minipage}
        &  
        \begin{minipage}[b]{0.12000\linewidth}
        \includegraphics[width=1\linewidth]{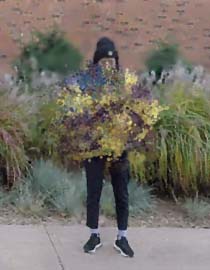} 
        \end{minipage}
        &  
        \begin{minipage}[b]{0.12000\linewidth}
        \includegraphics[width=1\linewidth]{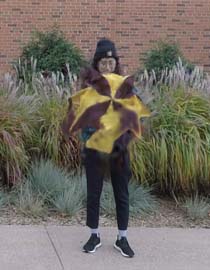} 
        \end{minipage}
        &  
        \begin{minipage}[b]{0.12000\linewidth}
        \includegraphics[width=1\linewidth]{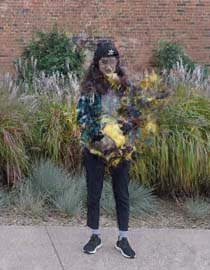} 
        \end{minipage}
        &  
        \begin{minipage}[b]{0.12000\linewidth}
        \includegraphics[width=1\linewidth]{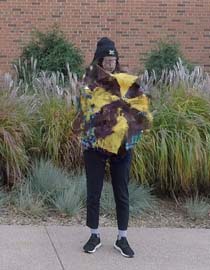} 
        \end{minipage}
        &  
        \begin{minipage}[b]{0.12000\linewidth}
        \includegraphics[width=1\linewidth]{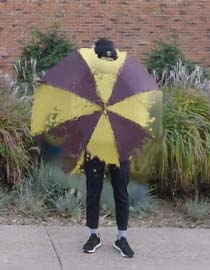} 
        \end{minipage}
        &  
        \begin{minipage}[b]{0.12000\linewidth}
        \includegraphics[width=1\linewidth]{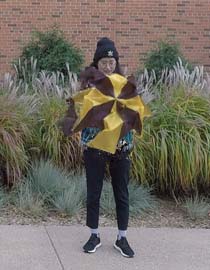}
        \end{minipage}
        &  
        \begin{minipage}[b]{0.12000\linewidth}
        \includegraphics[width=1\linewidth]{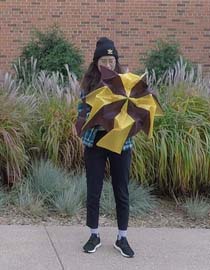}
        \end{minipage}
        \\ 
        \makebox[0.12000\linewidth][c]{\footnotesize MarbleGS~\cite{marblegs}} & 
        \makebox[0.12000\linewidth][c]{\footnotesize SoM~\cite{som} } & 
        \makebox[0.12000\linewidth][c]{\footnotesize SplineGS~\cite{splinegs}} & 
        \makebox[0.12000\linewidth][c]{\footnotesize MoDec-GS~\cite{modec-gs}} & 
        \makebox[0.12000\linewidth][c]{\footnotesize MoSca~\cite{mosca}} & 
        \makebox[0.12000\linewidth][c]{\footnotesize HiMoR~\cite{himor}} & 
        \makebox[0.12000\linewidth][c]{\footnotesize Ours} & 
        \makebox[0.12000\linewidth][c]{\footnotesize GT} \\
    \end{tabular}
    \vspace{-2mm}
    \caption{\textbf{Visual comparison of novel view synthesis on the NVIDIA dataset.} Note, images are cropped to highlight dynamic regions.}
    \label{fig:comnvidia}
\end{figure*}

\subsection{Dynamic Scene Reconstruction}
\label{sec:scenerecon}

\noindent \textbf{Scene initialization.}
\zxk{To avoid introducing deformation interference in static regions of the scene and reduce the computational overhead of deformation, we construct the scene using both static and dynamic Gaussians, ensuring that deformation calculations are performed only on the dynamic Gaussians. Specifically, we initialize static Gaussians by reprojecting the background depth across frames and share the same parameter composition as the original 3D Gaussians. Then, dynamic Gaussians are initialized by reprojecting the 2D tracklets' depth across frames. Dynamic Gaussians include an additional reference time parameter $t_{ref} \in \mathbb{R}$, which represents the timestamp of the depth map that initializes the dynamic Gaussian.}

\vspace{0.5em}
\noindent \textbf{Dynamic Gaussian deformation.}
\zxk{In our method, the dynamic Gaussians are deformed by following the deformation of their nearest SE(3) B-spline motion base. Specifically, for the dynamic Gaussian $g$, we first use the nearest-neighbor algorithm to obtain its nearest SE(3) B-spline motion base $T(t)^g$ at the reference time $t_{ref}^g$. Then, we use the k-nearest neighbors algorithm to obtain the $K$ SE(3) B-spline motion bases that are the nearest neighbors of $T(t)^g$ at all timestamps. Then, we use the DQB \cite{dqb} to fuse the relative pose transformation $\Delta Q$ of the SE(3) B-spline motion bases between the reference space state $Q_{ref}$ and the observation space state $Q_{obs}$ to obtain the relative pose transformation $\Delta Q^g$ of the dynamic Gaussian:}
\begin{equation}
    \Delta Q^g= DQB(\{(w_i,\Delta Q^i)\}^{K}_{i=1}), 
  \label{eq:motionfusion}
\end{equation}
\zxk{where $\Delta Q^i$ represents the relative pose transformation of the $i$-th SE(3) B-spline motion base $T(t)^i$ from the reference space to the observation space. The corresponding weight $w_i$ follows the weighting strategy in MoSca \cite{mosca}. Then, we calculate the position $\mu_g'$ and orientation $R_g'$ of the dynamic Gaussian $g$ in the observation space as follows:}
\begin{equation}
     \mu_g' = \Delta Q^g \mu_g, ~~~ R_g' = \Delta Q^g R_g.
  \label{eq:deformation}
\end{equation}

\noindent \textbf{Soft segment reconstruction.}
\zxk{To effectively model the temporary appearance and disappearance of dynamic objects in the scene, we use Eq.~(\ref{eq:motionfusion}) and Eq.~(\ref{eq:deformation}) to convert the dynamic Gaussians at all reference timestamps to the $i$-th observation timestamp. However, for monocular videos with very long frame rates, the time interval between the reference timestamp and the observation timestamp can affect the accuracy of the dynamic Gaussian transformations. To alleviate the severe interference caused by dynamic Gaussian transformations with long time intervals, we design a soft segment reconstruction strategy. This strategy ensures that the dynamic Gaussian at any observation timestamp is primarily composed of transformations from nearby local reference timestamps. We adjust the opacity $o$ of the dynamic Gaussians as follows:}
\begin{equation}
     o' = sigmoid(scale · (1- \lvert t_{ref}-t_{obs} \rvert)) * o,
  \label{eq:softsegment}
\end{equation}
\zxk{where $o'$ denotes the adjusted dynamic Gaussian opacity, and the scale is set to 5.0. Obviously, the opacity of the dynamic Gaussian at the reference timestamp with a longer interval from the observation timestamp is lower, thereby reducing the reconstruction uncertainty caused by the dynamic Gaussian transformations with a longer time interval.}

\subsection{Diffusion-based Multiview Prior}
\label{sec:multiviewdiffusion}

\zxk{Due to lack of multi-view cues in monocular videos, current dynamic scene reconstruction methods tend to overfit to the training views. This often leads to severe artifacts and motion blur in novel views, especially in regions with significant motion. Although traditional methods, such as background warping, can generate a target view image from a source view image to obtain multi-view information, the generated image only contains regions that are co-visible between the source and target views. Therefore, the multi-view information obtained by background warping does not include the invisible areas occluded by the training views.}

\zxk{To optimize the invisible areas in training views, we propose to use the scene prior information learned from the existing diffusion model to optimize the invisible areas. Specifically, we use the SDS loss to extract the prior knowledge in the diffusion model. To ensure the consistency of multi-view, we use a multi-view diffusion model \cite{zero123}, which can infer the target view image based on the source view image and the relative camera transformation between the source view and the target view. The SDS loss calculation process is as follows:}
\begin{equation}
    \mathcal{L}_{sds} = \mathbb{E}_{t, \boldsymbol{\epsilon}} \bigg[ \Vert \omega(t) ({\boldsymbol{\epsilon}_{\phi}}(z_t,t,P_{t}P_{s}^{-1},I_{s}) - {\boldsymbol{\epsilon}}) \Vert_2^2 \bigg],
  \label{eq:sdsloss}
\end{equation}
\zxk{where $P_s$ and $P_t$ denote the camera extrinsics of the source and target view, respectively. $I_s$ denotes the ground-truth image of the source view.} 

\subsection{Loss Function}
\label{sec:loss}
\noindent \textbf{Reconstruction loss.} \zxk{We supervise the training process with a reconstruction loss to align per-frame pixel-wise color input $\mathbf{I}$. The rendered image $\hat{\mathbf{I}}$ is supervised by the following reconstruction loss:}
\begin{equation}
    \mathcal{L}_{rec} = (1 - \beta)\mathcal{L}_{1}(\hat{\mathbf{I}}, \mathbf{I}) + \beta\mathcal{L}_{ssim}(\hat{\mathbf{I}}, \mathbf{I}),
  \label{eq:recloss}
\end{equation}
\zxk{where $\mathcal{L}_{1}$ and $\mathcal{L}_{ssim}$ are $L_1$ loss and SSIM \cite{ssim} loss, and $\beta=0.2$. We also constrain the scene geometry using the depth $\hat D$:}
\begin{equation}
    \mathcal{L}_{geo} = \mathcal{L}_{1}(\hat D, D).
  \label{eq:geoloss}
\end{equation}

\noindent \textbf{Motion smoothness loss.}
\zxk{Similar to MoSca \cite{mosca} and SoM \cite{som}, we introduce two motion smoothness losses, $\mathcal{L}_{arap}$ and $\mathcal{L}_{track}$, to enhance the smoothness of motion representation. $\mathcal{L}_{arap}$ is a multi-scale ARAP loss to constrain the motion bases to maintain locally rigid transformation as much as possible, and $\mathcal{L}_{track}$ leverages the optical flow estimated from training images to constrain the deformation of the dynamic Gaussians. Further details of the two losses are provided in the supplementary material.}

\vspace{0.5em}
\noindent \textbf{Camera smoothness loss.}
\zxk{Camera parameters in most monocular video datasets are often inaccurate. As a result, the scene cannot remain consistent across views, causing artifacts in the rendered images. To alleviate the impact of inaccurate camera parameters on scene reconstruction, we fine-tune the camera extrinsics $P$. Specifically, during the training process, we set the camera extrinsics as learnable parameters and optimize them together with the scene Gaussians and SE(3) B-spline motion bases. Considering that camera poses in monocular videos typically change smoothly over time, we design a camera smoothness loss $\mathcal{L}_{smo}$ to enforce temporal consistency of camera poses:}
\begin{equation}
    \mathcal{L}_{smo} = \textstyle \sum^{N_T-1}_{t=0} \big\|P_t^{-1}P_{t+1}\big\|^2_2 ,
  \label{eq:smoloss}
\end{equation}
where $P_t$ and $P_{t+1}$ denote the camera extrinsic of the $i$-th and $(i+1)$-th timestamps. The overall loss function for our model is:
\begin{equation}
\begin{split}
\mathcal{L} = & \ \lambda_{rec}\mathcal{L}_{rec} 
+ \lambda_{geo}\mathcal{L}_{geo} 
+ \lambda_{sds}\mathcal{L}_{sds} + \\
& \lambda_{arap}\mathcal{L}_{arap} 
+ \lambda_{track}\mathcal{L}_{track} 
+ \lambda_{smo}\mathcal{L}_{smo},
\end{split}
\label{eq:loss}
\end{equation}
\zxk{where we set $\lambda_{geo}=0.075$, $\lambda_{rec}=1.0$, $\lambda_{arap}=1.0$, $\lambda_{track}=1.0$, $\lambda_{sds}=0.01$, and $\lambda_{smo}=0.01$.}

\begin{figure*}[!t]
    \centering
    \setlength{\tabcolsep}{1.0pt}
    \scriptsize
    \begin{tabular}{cccccc}
        \begin{minipage}[b]{0.16200\linewidth}
        \includegraphics[width=1\linewidth]{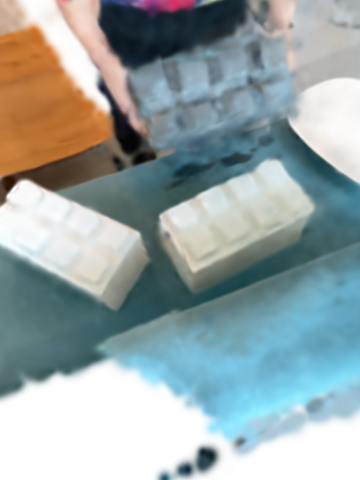} 
        \end{minipage}
        &  
        \begin{minipage}[b]{0.16200\linewidth}
        \includegraphics[width=1\linewidth]{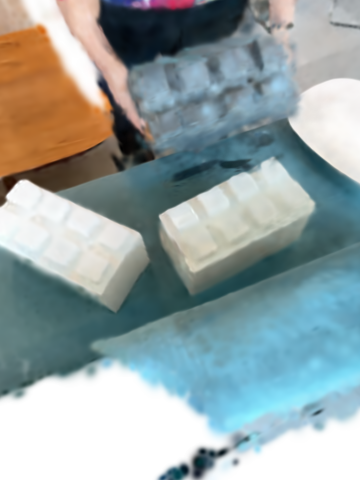} 
        \end{minipage}
        &  
        \begin{minipage}[b]{0.16200\linewidth}
        \includegraphics[width=1\linewidth]{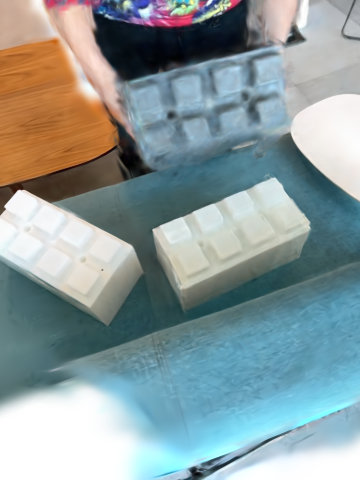} 
        \end{minipage}
        &  
        \begin{minipage}[b]{0.16200\linewidth}
        \includegraphics[width=1\linewidth]{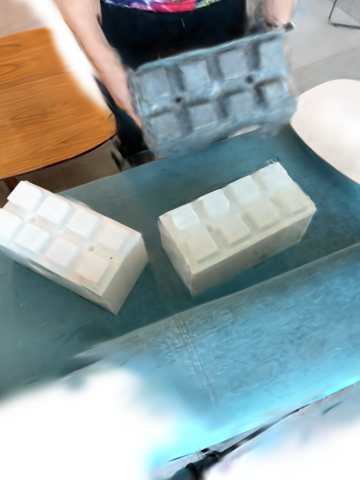}
        \end{minipage}
        &  
        \begin{minipage}[b]{0.16200\linewidth}
        \includegraphics[width=1\linewidth]{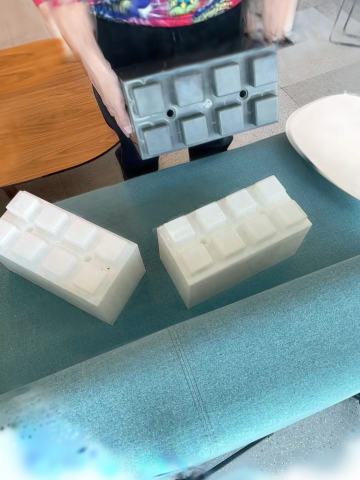}
        \end{minipage}
        & 
        \begin{minipage}[b]{0.16200\linewidth}
        \includegraphics[width=1\linewidth]{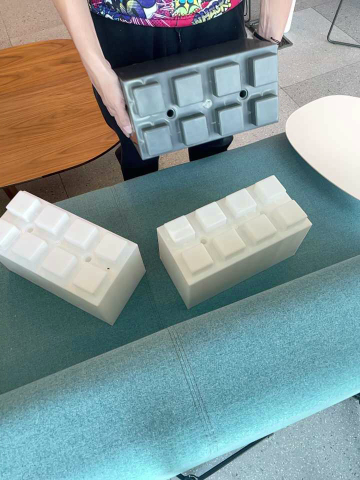}
        \end{minipage}
        \\
        \begin{minipage}[b]{0.16200\linewidth}
        \includegraphics[width=1\linewidth]{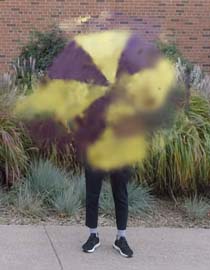} 
        \end{minipage}
        &  
        \begin{minipage}[b]{0.16200\linewidth}
        \includegraphics[width=1\linewidth]{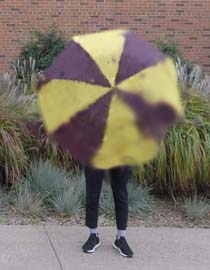} 
        \end{minipage}
        &  
        \begin{minipage}[b]{0.16200\linewidth}
        \includegraphics[width=1\linewidth]{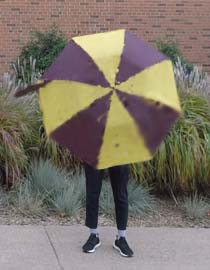} 
        \end{minipage}
        &  
        \begin{minipage}[b]{0.16200\linewidth}
        \includegraphics[width=1\linewidth]{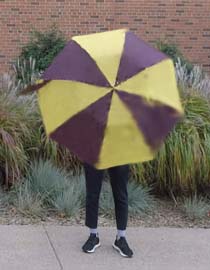}
        \end{minipage}
        &          
        \begin{minipage}[b]{0.16200\linewidth}
        \includegraphics[width=1\linewidth]{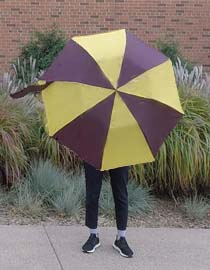}
        \end{minipage}
        &  
        \begin{minipage}[b]{0.16200\linewidth}
        \includegraphics[width=1\linewidth]{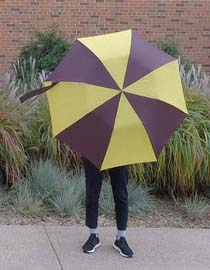}
        \end{minipage}
        \\ 
        \makebox[0.16200\linewidth][c]{\footnotesize w/o Adap.} & 
        \makebox[0.16200\linewidth][c]{\footnotesize w/o Soft.} & 
        \makebox[0.16200\linewidth][c]{\footnotesize w/o $\mathcal{L}_{sds}$} & 
        \makebox[0.16200\linewidth][c]{\footnotesize w/o $\mathcal{L}_{smo}$} & 
        \makebox[0.16200\linewidth][c]{\footnotesize Ours} & 
        \makebox[0.16200\linewidth][c]{\footnotesize GT} \\
    \end{tabular}
    \vspace{-2mm}
    \caption{\textbf{Visual ablation study on scenes from the iPhone and NVIDIA datasets.}}
    \label{fig:abla}
    \vspace{-10pt}
\end{figure*}

\begin{table*}[t]
    \centering
  
    \caption{\textbf{Quantitative comparison of novel view synthesis on the iPhone \cite{iphone} and  NVIDIA  \cite{nvidia} datasets.} Note that all training times are measured on the skating scene using a single NVIDIA RTX 4090 GPU, and FPS is measured on the same GPU at 480 × 360 resolution.}
    \vspace{-2mm}
    \scriptsize
    \resizebox{\linewidth}{!}
    {
    \begin{tabular}{l c c c c c c c c c}
    \toprule[0.5pt]
    \multirow{2}{*}{Method} & \multicolumn{3}{c}{iPhone} & \multicolumn{3}{c}{NVIDIA}  &\multirow{2}{*}{Param.} & \multirow{2}{*}{Training Time} & \multirow{2}{*}{FPS}\\ 
    & mPSNR$\uparrow$ & mSSIM$\uparrow$ & mLPIPS$\downarrow$ & PSNR$\uparrow$ & SSIM$\uparrow$ & LPIPS$\downarrow$ \\ 
    \midrule

    MarbleGS~\cite{marblegs} & 16.08 & 0.568 & 0.433 & 24.56 & 0.667 & 0.110 & 51.6M & 13 hrs & 8.367\\
    SoM~\cite{som} & 17.13 & 0.674 & 0.279 & 24.58 & 0.651 & 0.124 & 173.4M & 1 hr & 38.754\\
    HiMoR~\cite{himor} & 16.02 & 0.558 & 0.325 & 24.89 & 0.694 & 0.118 & 185.2M & 1 hr & 34.156\\
    MoDec-GS~\cite{modec-gs} & 14.65 & 0.320 & 0.461 & 23.94 & 0.628 & 0.132 & 86.1M & 1 hr & 9.074\\
    SplineGS~\cite{splinegs} & 15.52 & 0.483 & 0.371 & 27.12 & 0.872 & 0.052 & 158.4M & 2 hrs & 387.469\\
    MoSca~\cite{mosca} & 19.33 & 0.718 & 0.274 & 26.76 & 0.854 & 0.070 & 98.5M & 30 mins & 35.425 \\
    Ours &\textbf{20.17} & \textbf{0.729} & \textbf{0.274} & \textbf{27.81} & \textbf{0.871} & \textbf{0.049} & 134.7M & 30 mins & 45.124 \\    
    
    \bottomrule[0.5pt]
    \end{tabular}
    }
\label{tab:com}
\vspace{-10pt}
\end{table*}

\begin{table}[!t]
    \centering
    \caption{\textbf{Ablation studies on the iPhone and NVIDIA dataset.} ``w/o Adap.'' denotes removing the adaptive control in SE(3) B-spline motion bases, ``w/o Soft.'' means omitting soft segment reconstruction, and ``w/o $\mathcal{L}_{sds}$'' and ``w/o $\mathcal{L}_{smo}$'' refer to ignoring multi-view SDS loss and the camera smoothness loss, respectively.}
     \vspace{-2mm}
    \footnotesize
    \setlength{\tabcolsep}{3pt} 
    {
    \begin{tabular}{l c c c c c c}
    \toprule[0.5pt]
    \multirow{2}{*}{Method} & \multicolumn{3}{c}{iPhone} & \multicolumn{3}{c}{NVIDIA}  \\ 
    & mPSNR$\uparrow$ & mSSIM$\uparrow$ & mLPIPS$\downarrow$ & PSNR$\uparrow$ & SSIM$\uparrow$ & LPIPS$\downarrow$ \\ 
    \midrule

    w/o Adap.& 18.84 & 0.685 & 0.350 & 26.87 & 0.803 & 0.128 \\
    w/o Soft. & 19.02 & 0.628 & 0.328 & 27.06 & 0.832 & 0.085 \\
    w/o $\mathcal{L}_{sds}$ & 19.39 & 0.715 & 0.288 & 27.13 & 0.845 & 0.074 \\
    w/o $\mathcal{L}_{smo}$ & 19.18 & 0.713 & 0.295 & 27.15 & 0.861 & 0.076 \\
    Ours &\textbf{20.17} & \textbf{0.729} & \textbf{0.274} & \textbf{27.81} & \textbf{0.871} & \textbf{0.049} \\  
    \bottomrule[0.5pt]
    \end{tabular}
    }
\label{tab:abla}
\vspace{-10pt}
\end{table}

\subsection{Implementation Details}

\zxk{We set $N_{prune} =500$, $N_{densify}=500$, and $N_c=100$. We employ the Adam optimizer \cite{adam} to jointly optimize the Gaussians, the SE(3) B-spline motion bases, and the camera extrinsic parameters. The learning rates are set to $1.6 \times 10^{-4}$ for SE(3) B-spline motion bases, and $3 \times 10^{-4}$ for camera extrinsic parameters. We train each scene for 8,000 iterations. More implementation details are given in the supplementary material.}


\section{Experiments}
\label{sec:exper}

\noindent \textbf{Datasets.} We evaluate our method on two datasets, \ie iPhone~\cite{iphone} and NVIDIA~\cite{nvidia}. The iPhone dataset contains generic, diverse dynamic scenes captured with a handheld iPhone using realistic camera motions for training, and utilizes two static cameras at significantly different poses from the training views for testing. It comprises 14 scenes, 7 of which contain multi-camera captures for the evaluation of novel view synthesis. Following SoM \cite{som}, we use five of these scenes and exclude two due to camera pose inaccuracies. The NVIDIA dataset consists of seven scenes captured with a rig with 12 cameras, where all scenes are used. 

\vspace{0.5em}
\noindent \textbf{Metrics.} Akin to previous work \cite{marblegs, som, mosca,himor}, we compute mPSNR, mSSIM \cite{ssim}, and mLPIPS \cite{lpips} over regions indicated by a co-visibility mask on the iPhone dataset. For the NVIDIA dataset, we directly compute PSNR, SSIM, and LPIPS for evaluation. We also adopt the percentage of correctly transferred keypoints (PCK-T) to evaluate long-range tracking from training views.

\vspace{0.5em}

\subsection{Comparison with State-of-the-Art Methods}
\noindent \textbf{Baselines.}
We compare our method with various state-of-the-art methods, including Shape-of-Motion \cite{som}, MoSca \cite{mosca}, Gaussian Marbles \cite{marblegs}, SplineGS \cite{splinegs}, MoDec-GS \cite{modec-gs}, and HiMoR \cite{himor}. For fair comparison, we produce their results using publicly available implementations or trained models provided by the authors with the recommended parameter settings.

\begin{table}[!t]
    \centering
    \small
    \caption{\textbf{Ablation study of SE(3) B-spline motion bases on the iPhone dataset.} ``w/ Pose trans.'' means replacing our motion representation with the SE(3)-based pose transformation used in SoM \cite{som}. ``w/ Motion scaff.'' indicates replacing the representation with the motion scaffolds used in MoSca \cite{mosca}.}
    \vspace{-2mm}
    \begin{tabular}{lccc}
        \toprule
        Method & mPSNR$\uparrow$ & mSSIM$\uparrow$ & mLPIPS$\downarrow$ \\
        \midrule
        w/ Pose trans.  & 18.17 & 0.675 & 0.328 \\
        w/ Motion scaff.  & 19.26 & 0.698 & 0.296 \\
        Ours  & \textbf{20.17} & \textbf{0.729} & \textbf{0.274} \\
        \bottomrule
    \end{tabular}
    \label{tab:ablase3}
    \vspace{-10pt}
\end{table}

\vspace{0.5em}
\noindent \textbf{Evaluation on novel view synthesis.}
Table \ref{tab:com} reports the quantitative results, where we can see that our method outperforms others on both the two datasets. Figures \ref{fig:comiphone} and \ref{fig:comnvidia} further present visual comparisons of novel view synthesis. As shown, our method produces better novel view synthesis results for both dynamic and static regions, while the compared methods struggle to provide sharp results with well-preserved structural details. Please see the supplementary material for more comparison results. 

\begin{figure}[!t]
    \centering
    \setlength{\tabcolsep}{1.0pt}
    \scriptsize
    \begin{tabular}{cccc}
        \begin{minipage}[b]{0.24300\columnwidth}
        \includegraphics[width=1\linewidth]{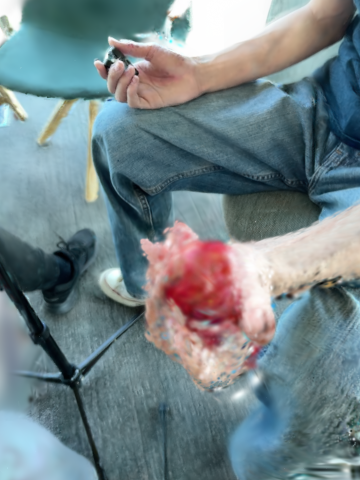} 
        \end{minipage}
        &  
        \begin{minipage}[b]{0.24300\columnwidth}
        \includegraphics[width=1\linewidth]{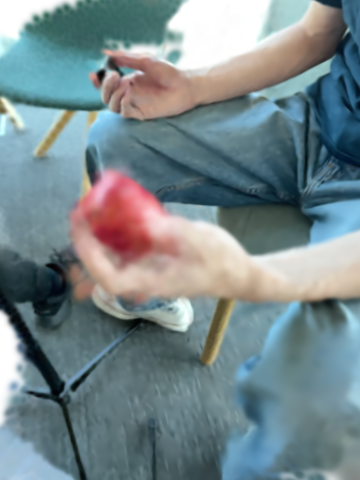} 
        \end{minipage}
        &  
        \begin{minipage}[b]{0.24300\columnwidth}
        \includegraphics[width=1\linewidth]{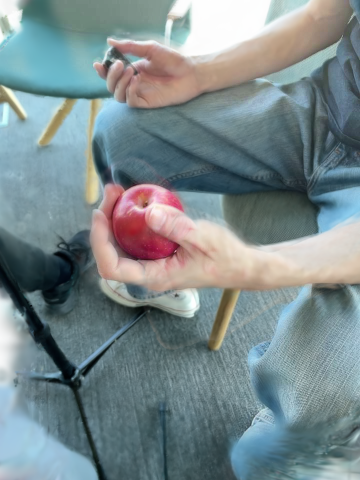} 
        \end{minipage}
        &  
        \begin{minipage}[b]{0.24300\columnwidth}
        \includegraphics[width=1\linewidth]{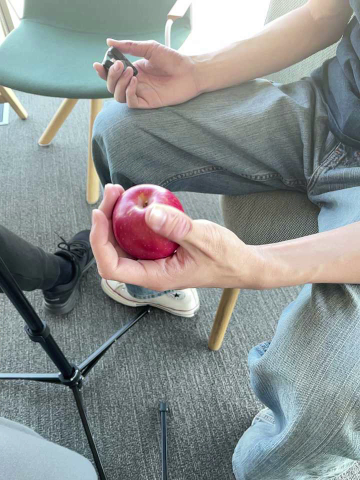}
        \end{minipage}
        \\ 
        \begin{minipage}[b]{0.24300\columnwidth}
        \includegraphics[width=1\linewidth]{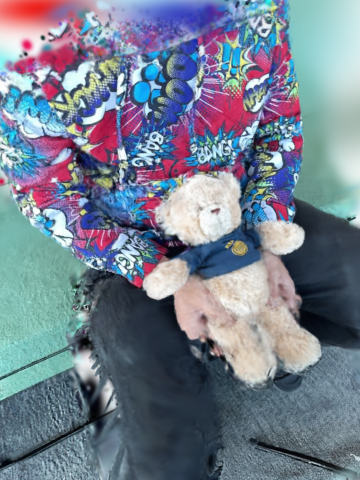} 
        \end{minipage}
        &  
        \begin{minipage}[b]{0.24300\columnwidth}
        \includegraphics[width=1\linewidth]{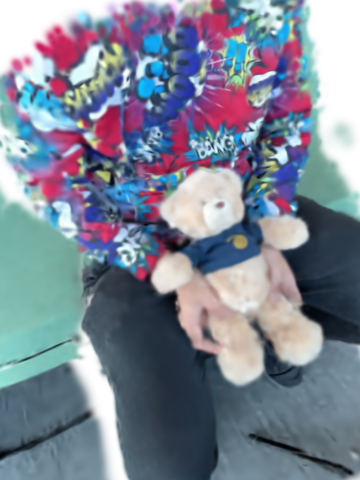} 
        \end{minipage}
        &  
        \begin{minipage}[b]{0.24300\columnwidth}
        \includegraphics[width=1\linewidth]{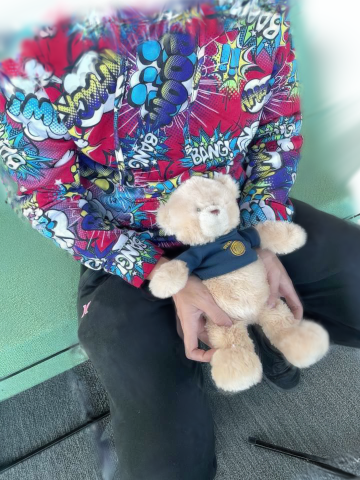} 
        \end{minipage}
        &  
        \begin{minipage}[b]{0.24300\columnwidth}
        \includegraphics[width=1\linewidth]{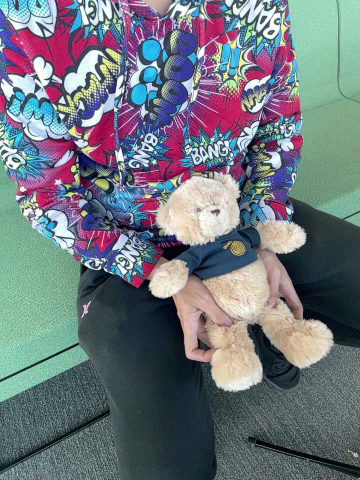}
        \end{minipage}
        \\ 
        \makebox[0.24300\columnwidth][c]{\footnotesize w/ Pose trans.} & 
        \makebox[0.24300\columnwidth][c]{\footnotesize w/ Motion scaff.} & 
        \makebox[0.24300\columnwidth][c]{\footnotesize Ours} &
        \makebox[0.24300\columnwidth][c]{\footnotesize GT} \\
    \end{tabular}
    \vspace{-2mm}
    \caption{\textbf{Effect of SE(3) B-spline Motion Bases.}}
    \label{fig:ablase3}
    \vspace{-10pt}
\end{figure}

\vspace{0.5em}
\noindent \textbf{Evaluation on correspondence.}
We also quantitatively evaluate the correspondence tracking accuracy on the iPhone dataset. Comparing the results in Table \ref{tab:correspondence}, it is clear that our method outperforms other methods, demonstrating our superiority in obtaining accurate correspondence. 


\subsection{More Analysis}

\noindent \textbf{Ablation studies.}
Here we conduct ablation studies to evaluate the contribution of each component in our model, \ie, adaptive control of motion bases, soft segment reconstruction, the multi-view SDS loss, and the camera smooth loss. As shown by the numerical results in Table \ref{tab:abla}, each of our design has a clear contribution to our performance. Besides, we in Figure \ref{fig:abla} qualitatively validate the necessity of each component in our model. We also assess the effectiveness of SE(3) B-spline motion bases in Figure \ref{fig:ablase3} and Table \ref{tab:ablase3}. Specifically, we compare it with two alternatives: (i) SE(3)-based pose transformation in SoM \cite{som} (w/ Pose trans.), (ii) motion scaffolds in MoSca \cite{mosca} (w/ Motion scaff.). As shown, our explicit continuous motion representation based on SE(3) B-spline motion bases clearly outperforms the two alternatives, manifesting its effectiveness. Please see the supplementary material for more ablation results. 

\vspace{0.5em}
\noindent \textbf{Effect of 2D prior errors.}
Similar to previous methods \cite{som,marblegs,mosca, splinegs}, our method also leverages a 2D tracking prior from pretrained models \cite{bootstap}. Unlike their sensitivity to error of 2D tracking prior, our method has some tolerance to the error due to our explicit continuous motion representation. Specifically, the continuous nature of SE(3) B-spline motion bases can help mitigate tracking jitter error in 2D tracking prior. We also quantitatively evaluate how errors from external 2D tracking prior affect the performance of our method. As shown by the numerical results in Table \ref{tab:prior}, the performance drop led by a random perturbation of 2D tracking prior is quite limited, validating our robustness to 2D prior errors.



\begin{figure}[!t]
    \centering
    \setlength{\tabcolsep}{1.0pt}
    \scriptsize
    \begin{tabular}{cccc}
        \begin{minipage}[b]{0.24300\columnwidth}
        \includegraphics[width=1\linewidth]{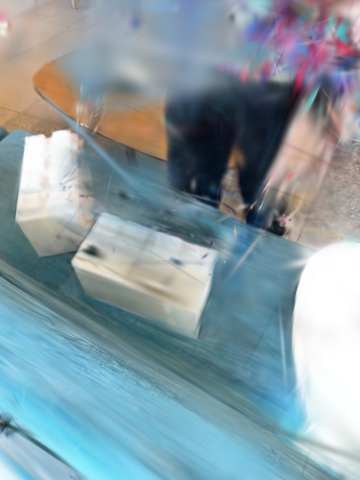} 
        \end{minipage}
        &  
        \begin{minipage}[b]{0.24300\columnwidth}
        \includegraphics[width=1\linewidth]{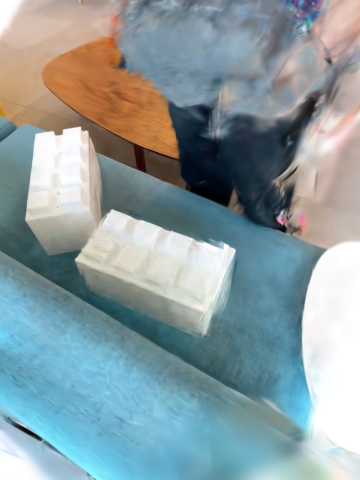} 
        \end{minipage}
        &  
        \begin{minipage}[b]{0.24300\columnwidth}
        \includegraphics[width=1\linewidth]{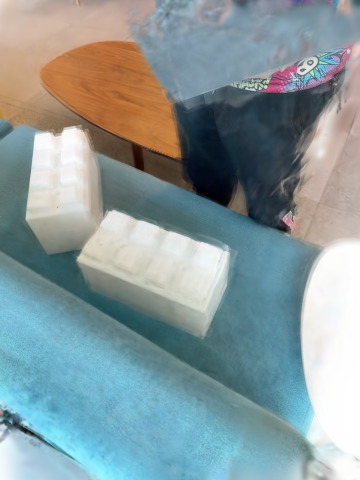} 
        \end{minipage}
        &  
        \begin{minipage}[b]{0.24300\columnwidth}
        \includegraphics[width=1\linewidth]{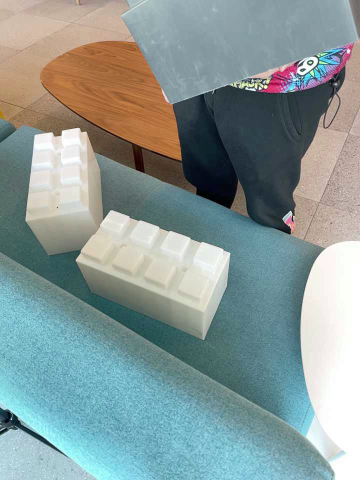}
        \end{minipage}
        \\ 
        \begin{minipage}[b]{0.24300\columnwidth}
        \includegraphics[width=1\linewidth]{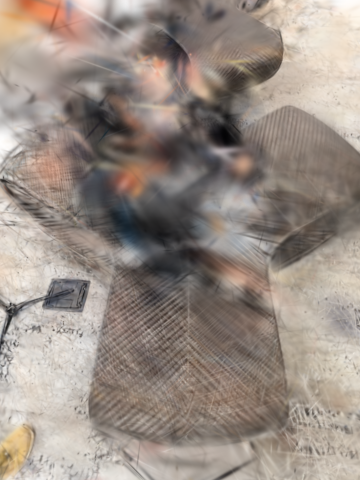} 
        \end{minipage}
        &  
        \begin{minipage}[b]{0.24300\columnwidth}
        \includegraphics[width=1\linewidth]{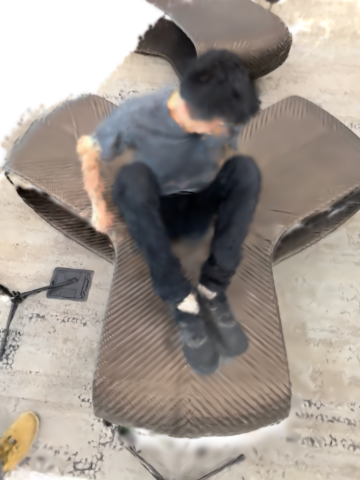} 
        \end{minipage}
        &  
        \begin{minipage}[b]{0.24300\columnwidth}
        \includegraphics[width=1\linewidth]{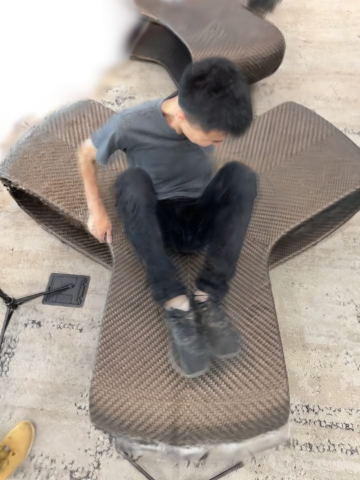} 
        \end{minipage}
        &  
        \begin{minipage}[b]{0.24300\columnwidth}
        \includegraphics[width=1\linewidth]{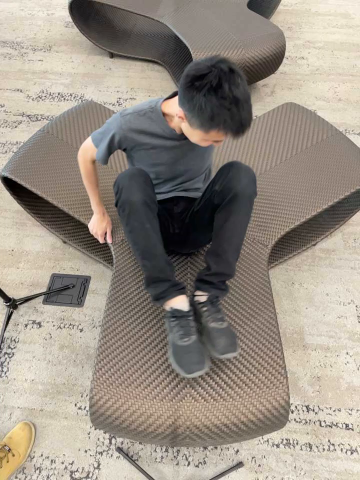}
        \end{minipage}
        \\ 
        \makebox[0.24300\columnwidth][c]{\footnotesize MoDec-GS~\cite{modec-gs}} & 
        \makebox[0.24300\columnwidth][c]{\footnotesize MoSca~\cite{mosca}} & 
        \makebox[0.24300\columnwidth][c]{\footnotesize Ours} &
        \makebox[0.24300\columnwidth][c]{\footnotesize GT} \\
    \end{tabular}
    \vspace{-2mm}
    \caption{\textbf{Failure cases.} Our method fails to handle dynamic scenes with large non-rigid motion.}
    \label{fig:failcase}
    \vspace{-5pt}
\end{figure}

\begin{table}[!t]
    \centering
    \small
    \setlength{\tabcolsep}{4.0pt}
    \renewcommand\arraystretch{0.8}
    \caption{\textbf{Comparison of correspondence on the iPhone dataset.}}
    \vspace{-2mm}
    \resizebox{\linewidth}{!}{
    \begin{tabular}{lcccc}
        \toprule
        Metric & MarbleGS~\cite{marblegs} & HiMoR~\cite{himor} & MoSca~\cite{mosca} & Ours \\
        \midrule
        PCK-T$\uparrow$ & 0.803 & 0.809 & 0.821 & \textbf{0.833} \\
        \bottomrule
    \end{tabular}
    }
    \label{tab:correspondence}
\end{table} 


\begin{table}[!t]
    \centering
    \caption{\textbf{Effect of 2D prior errors.} ``w/ Prior pert.'' indicates that the 2D tracking prior is perturbed by adding random noise within the range of [-15, 15].}
     \vspace{-2mm}
    \resizebox{\linewidth}{!}{
    \setlength{\tabcolsep}{3pt} 
    {
    \begin{tabular}{l c c c c c c}
    \toprule[0.5pt]
    \multirow{2}{*}{Method} & \multicolumn{3}{c}{iPhone} & \multicolumn{3}{c}{NVIDIA}  \\ 
    & mPSNR$\uparrow$ & mSSIM$\uparrow$ & mLPIPS$\downarrow$ & PSNR$\uparrow$ & SSIM$\uparrow$ & LPIPS$\downarrow$ \\ 
    \midrule

    w/ Prior pert. & 20.11 & 0.725 & 0.278 & 27.74 & 0.867 & 0.051 \\
    Ours &\textbf{20.17} & \textbf{0.729} & \textbf{0.274} & \textbf{27.81} & \textbf{0.871} & \textbf{0.049} \\  
    \bottomrule[0.5pt]
    \end{tabular}
    }}
\label{tab:prior}
\vspace{-10pt}
\end{table}

\vspace{0.5em}
\noindent \textbf{Limitations.} Our method has limitations.
First, as shown in Figure \ref{fig:failcase}, dynamic scenes with substantial non-rigid deformation and motion blur remain a challenge for our method. Second, our method fails to handle blurry monocular videos with rapid camera or object motion. 

\section{Conclusion}
\label{sec:conclu}

We have presented a framework that allows high-quality dynamic Gaussian Splatting from monocular videos. Our key idea is to explicitly model the continuous Gaussian position and orientation deformation trajectories. To this end, we devise an explicit continuous SE(3) B-spline motion bases, which control dynamic Gaussian trajectories using SE(3) Cumulative B-spline with a small number of control points. Besides, we introduce a soft segment reconstruction strategy to mitigate the interference of dynamic Gaussians induced by long-time interval motion deformation on scene reconstruction. Furthermore, we propose to augment scene multi-view cues by using a multi-view diffusion model with rich priors to introduce SDS loss on the invisible regions under the training view in monocular videos. Extensive experiments demonstrate the superiority of our method.

\vspace{0.5em}
\noindent \textbf{Acknowledgement.} This work was supported by the National Natural Science Foundation of China (62471499), the Guangdong Basic and Applied Basic Research Foundation (2023A1515030002).

{
    \small
    \bibliographystyle{ieeenat_fullname}
    \bibliography{main}
}

\clearpage
\clearpage
\setcounter{page}{1}
\maketitlesupplementary

\section{Additional Details}
\noindent \textbf{Initiation of motion bases.}
We use the 2D tracklets $\tau$ obtained from the pre-trained model \cite{bootstap} to initialize the motion bases.
\begin{equation}
  \tau = [(p_t,v_t)]_{t=1}^{N_T},
  \label{eq:tracklet}
\end{equation}
where $p_t$ and $v_t$ represent the position and visibility of the 2D tracklet, respectively. We lift the 2D tracklet position into 3D using depth reprojection. For training view $v$, given the 2D tracklet $p_v$, we obtain the corresponding 3D position $q_v$ as follows:
\begin{equation}
  q_v = P_v\pi_K^{-1}(p_v,D_v[p_v]),
  \label{eq:depthreproject}
\end{equation}
where $\pi_K^{-1}$ denotes the back-projection using the camera intrinsics $K$, $P_v$ denotes the camera extrinsics of training view $v$, and $D_v$ represents the depth map of the same view. Since the depth values at the locations corresponding to invisible 2D tracklets do not reflect the true depth after reprojection, these invisible tracklets degrade the initialization of the motion bases. To mitigate this issue, we replace the 3D reprojection values of invisible tracklets with those of the nearest visible 3D position $q_v$ using linear interpolation.

Then, we initialize the orientation of all 3D points with the identity matrix $I$. Finally, we uniformly sample 3D points over time as control points for the SE(3) B-spline motion bases.

{\vspace{0.5em}
\noindent \textbf{Details of multi-view diffusion model.}
\zxkk{We use Zero123-xl-diffusers (Stable-Diffusion v1.5) as the multi-view diffusion model, where the conditioning inputs consist of a reference image and the corresponding relative camera pose. Following DreamScene4D \cite{dreamscene4d}, the reference image contains only the foreground region of the input. We achieve view sampling by applying a small random perturbation to the center of the training camera.

Zero123-xl-diffusers can only provide an object-centric multi-view SDS prior, a naive application of object-centric multi-view SDS prior will inevitably suffer from a domain gap issue. However, we mitigate this by the following operations: (i) similar to DreamScene4D \cite{dreamscene4d}, we apply SDS loss only over foreground objects, thereby reducing the scene-level task to a foreground object modeling problem. (ii) we perform view sampling near the training views to ensure a reliable diffusion prior. (iii) we use rendered and reference images from the same time frame to increase geometric consistency of foreground objects. }

\begin{figure}[!t]
    \centering
    \setlength{\tabcolsep}{1.0pt}
    \scriptsize
    \begin{tabular}{cccc}
        \begin{minipage}[b]{0.24300\columnwidth}
        \includegraphics[width=1\linewidth]{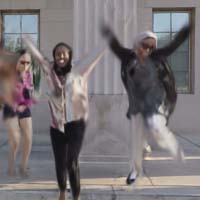} 
        \end{minipage}
        &  
        \begin{minipage}[b]{0.24300\columnwidth}
        \includegraphics[width=1\linewidth]{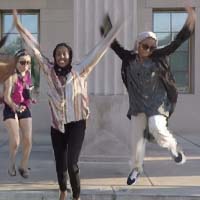} 
        \end{minipage}
        &  
        \begin{minipage}[b]{0.24300\columnwidth}
        \includegraphics[width=1\linewidth]{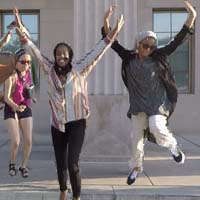} 
        \end{minipage}
        &  
        \begin{minipage}[b]{0.24300\columnwidth}
        \includegraphics[width=1\linewidth]{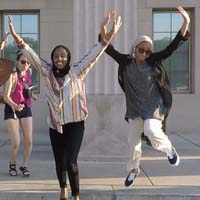}
        \end{minipage}
        \\ 
        \makebox[0.24300\columnwidth][c]{\footnotesize 4DGaussian} & 
        \makebox[0.24300\columnwidth][c]{\footnotesize MarbleGS} & 
        \makebox[0.24300\columnwidth][c]{\footnotesize Ours} &
        \makebox[0.24300\columnwidth][c]{\footnotesize GT} \\
    \end{tabular}
    \vspace{-2mm}
    \caption{\textbf{Visual comparison of novel view synthesis.}}
    \label{fig:com_perpoint_deformation_cr}
    \vspace{-5pt}
\end{figure}

\begin{figure}[!t]
    \centering
    \setlength{\tabcolsep}{1.0pt}
    \scriptsize
    \begin{tabular}{cccc}
        \begin{minipage}[b]{0.24300\columnwidth}
        \includegraphics[width=1\linewidth]{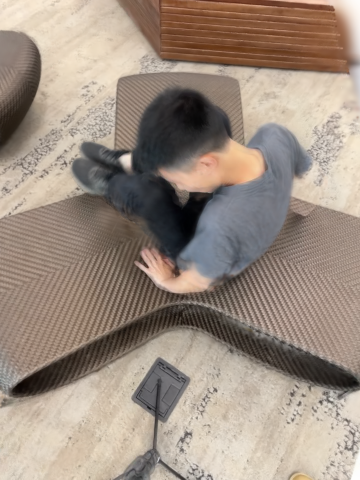} 
        \end{minipage}
        &  
        \begin{minipage}[b]{0.24300\columnwidth}
        \includegraphics[width=1\linewidth]{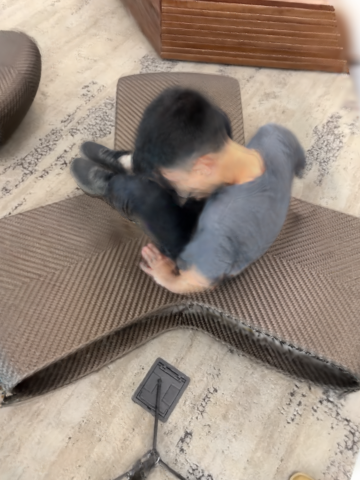} 
        \end{minipage}
        &  
        \begin{minipage}[b]{0.24300\columnwidth}
        \includegraphics[width=1\linewidth]{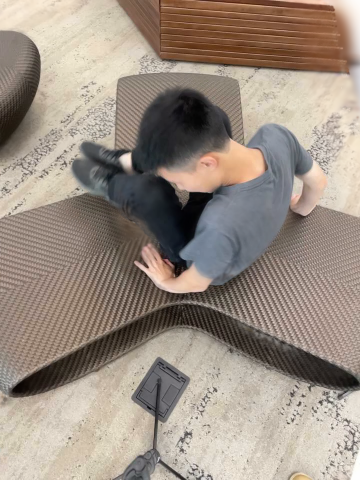} 
        \end{minipage}
        &  
        \begin{minipage}[b]{0.24300\columnwidth}
        \includegraphics[width=1\linewidth]{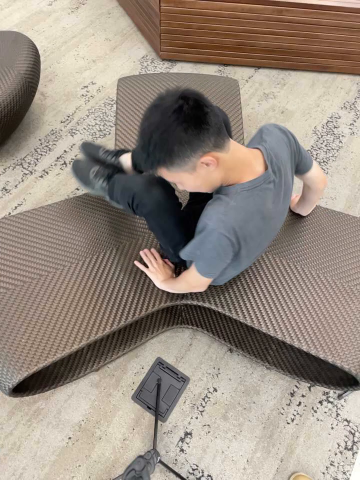}
        \end{minipage}
        \\ 
        \begin{minipage}[b]{0.24300\columnwidth}
        \includegraphics[width=1\linewidth]{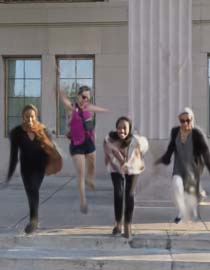} 
        \end{minipage}
        &  
        \begin{minipage}[b]{0.24300\columnwidth}
        \includegraphics[width=1\linewidth]{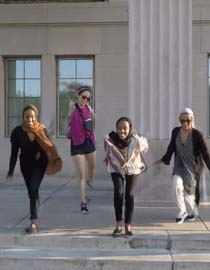} 
        \end{minipage}
        &  
        \begin{minipage}[b]{0.24300\columnwidth}
        \includegraphics[width=1\linewidth]{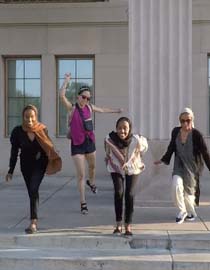} 
        \end{minipage}
        &  
        \begin{minipage}[b]{0.24300\columnwidth}
        \includegraphics[width=1\linewidth]{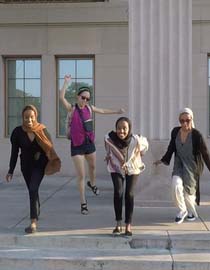}
        \end{minipage}
        \\ 
        \makebox[0.24300\columnwidth][c]{\footnotesize w/ Random} & 
        \makebox[0.24300\columnwidth][c]{\footnotesize w/ All} & 
        \makebox[0.24300\columnwidth][c]{\footnotesize Ours} &
        \makebox[0.24300\columnwidth][c]{\footnotesize GT} \\
    \end{tabular}
    \vspace{-2mm}
    \caption{\textbf{Effect of pruning strategy.}}
    \label{fig:ablaprune}
\end{figure}

\begin{table}[!t]
    \centering
     \vspace{-2mm}
    \caption{\textbf{Ablation on the iPhone and NVIDIA datasets.} ``w/ Random'' denotes randomly pruning a control point with error below $\epsilon_{prune}$, while ``w/ All'' prunes all control points with errors below $\epsilon_{prune}$.}
     \vspace{-2mm}
    \footnotesize
    \setlength{\tabcolsep}{3pt} 
    {
    \begin{tabular}{l c c c c c c}
    \toprule[0.5pt]
    \multirow{2}{*}{Method} & \multicolumn{3}{c}{iPhone} & \multicolumn{3}{c}{NVIDIA}  \\ 
    & mPSNR$\uparrow$ & mSSIM$\uparrow$ & mLPIPS$\downarrow$ & PSNR$\uparrow$ & SSIM$\uparrow$ & LPIPS$\downarrow$ \\ 
    \midrule
    w/ Random & 19.84 & 0.721 & 0.283 & 27.59 & 0.853 & 0.062 \\
    w/ All & 19.37 & 0.716 & 0.293 & 27.26 & 0.848 & 0.065 \\
    Ours &\textbf{20.17} & \textbf{0.729} & \textbf{0.274} & \textbf{27.81} & \textbf{0.871} & \textbf{0.049} \\  
    \bottomrule[0.5pt]
    \end{tabular}
    }
\label{tab:ablaprune}
\vspace{-10pt}
\end{table}

\noindent \textbf{Motion smoothness loss.}
Similar to MoSca \cite{mosca}, we maintain motion smoothness and propagate the visible information to the unknowns by optimizing a physics-inspired as-rigid-as-possible (ARAP) loss. Given two timestamps separated by a time interval $\Delta$, we define the ARAP loss $\mathcal{L}_{arap}$ as:
\begin{align}
    \mathcal{L}_{arap}&= \sum_{m=1}^{N_m} \sum_{n \in K(m)}\left | \| t_t^{(m)} - t_t^{(n)}\| - \| t_{t+\Delta}^{(m)} - t_{t+\Delta}^{(n)}\| \right | \nonumber
    \\
    &+ \left\| Q^{-1 \, (n)}_t  t^{(m)}_t - Q^{-1 \, (n)}_{t+\Delta} t^{(m)}_{t+\Delta} \right\|,
    \label{eq: loss_arap}
\end{align}
where $n \in K(m)$ denotes that motion base $n$ is one of the k-nearest neighbors of motion base $m$. The first term encourages the preservation of local distances within the neighborhood, while the second term preserves local coordinates by involving the local frame $Q$ in the optimization. Similar to SoM \cite{som}, we maintain dynamic Gaussian deformation smoothness by optimizing the track loss. Specifically, we additionally render the 2D tracks $\hat p_{t \to t'}$ for a pair of randomly sampled timestamps $t$ (query) and $t'$ (target). We supervise these rendered correspondences with the lifted long-range 2D track:
\begin{equation}
  \mathcal{L}_{track} = \big\|p_{t \to t'}-\hat p_{t \to t'}\big\|^2_2.
  \label{eq:depthreproject}
\end{equation}

\zxkk{ \section{Additional Comparison Results}
\noindent \textbf{Comparison with Per-Point Deformation Methods.}
As shown in Figure \ref{fig:com_perpoint_deformation_cr}, both our motion representation and per-point deformation field (PPDF) based methods (4DGaussians and MarbleGS) struggle to handle large non-rigid motion, but our method produces a slightly better result. The reason is that, our method allows us to fuse dynamic Gaussians across all frames to construct the target frame's dynamic foreground, while PPDF is highly prone to overfitting to the training view despite its strong representation flexibility of motion. }


\vspace{0.5em}
\noindent \textbf{More qualitative results.}
Figures \ref{fig:apple}, \ref{fig:paperwindmill}, \ref{fig:balloon1}, and \ref{fig:umbrella} provide more visual comparison of novel view synthesis results on the iPhone \cite{iphone} and NVIDIA \cite{nvidia} dataset.
As shown, our method clearly outperforms previous methods.

\section{Additional Ablation Results}
\noindent \textbf{Adaptive control mechanism.}
\zxkk{We conduct an ablation study to evaluate the effectiveness of our control point pruning strategy. Our method selects the control point with the smallest pruning error among those whose errors are below the threshold $\epsilon_{prune}$. We compare it with two alternatives: (i) randomly pruning one control point whose pruning error is below $\epsilon_{prune}$ (w/ Random), and (ii) pruning all control points whose pruning errors are below $\epsilon_{prune}$ (w/ All). As shown in Table \ref{tab:ablaprune}, our strategy achieves the best performance.}

\begin{figure}[!t]
    \centering
    \includegraphics[width=0.99\columnwidth]{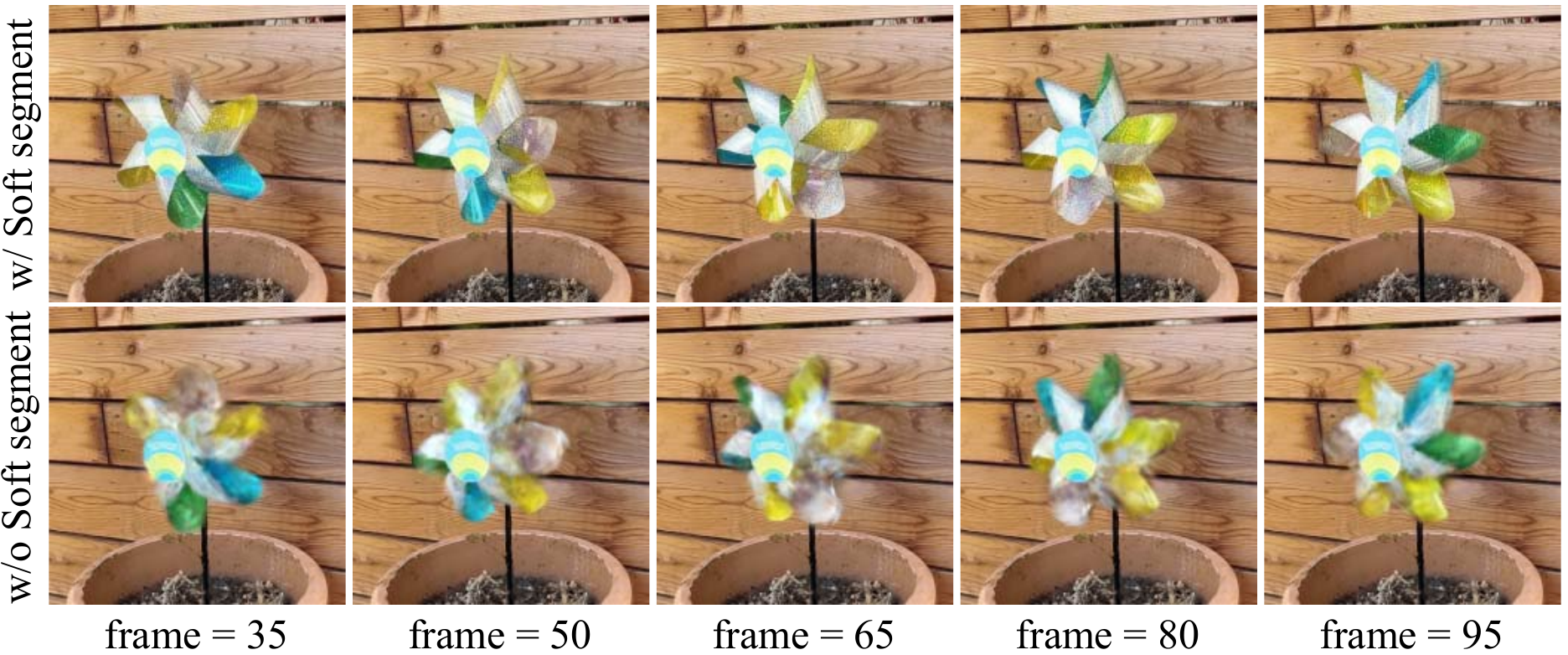}
    \vspace{-2mm}
    \caption{
        \textbf{Effect of soft segmentation reconstruction.} 
    }
    \label{fig:ablasoft_cr}
    \vspace{-6mm}
\end{figure}

\zxkk{\vspace{0.5em}
\noindent \textbf{Soft segment reconstruction.}
We conduct an ablation study to evaluate the effectiveness of our soft segment reconstruction. As shown in Figure \ref{fig:ablasoft_cr}, soft segment reconstruction helps to reconstruct dynamic paper-windmill (cropped view) with better long-term consistency.}

\begin{figure*}[!h]
    \centering
    \setlength{\tabcolsep}{1.0pt}
    \scriptsize
    \begin{tabular}{cccccc}
        \begin{minipage}[b]{0.16300\linewidth}
        \includegraphics[width=1\linewidth]{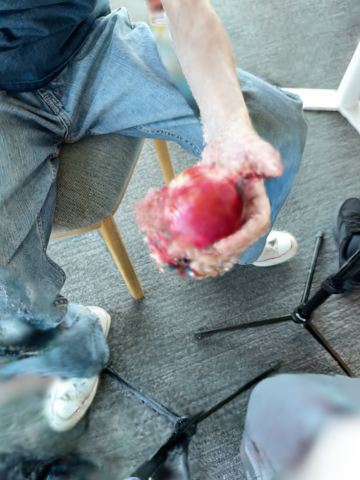}
        \end{minipage}
        &  
        \begin{minipage}[b]{0.16300\linewidth}
        \includegraphics[width=1\linewidth]{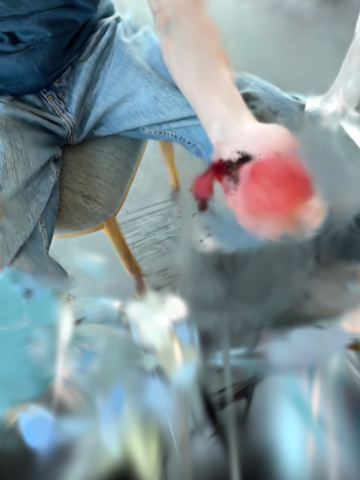} 
        \end{minipage}
        &  
 
        \begin{minipage}[b]{0.16300\linewidth}
        \includegraphics[width=1\linewidth]{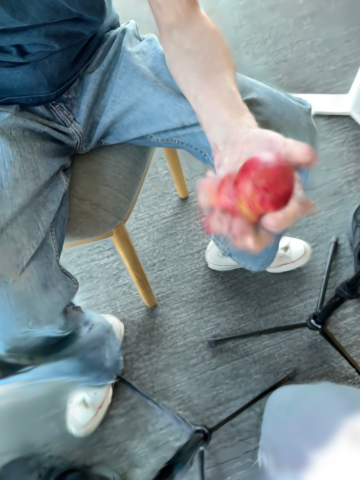} 
        \end{minipage}
        &  
        \begin{minipage}[b]{0.16300\linewidth}
        \includegraphics[width=1\linewidth]{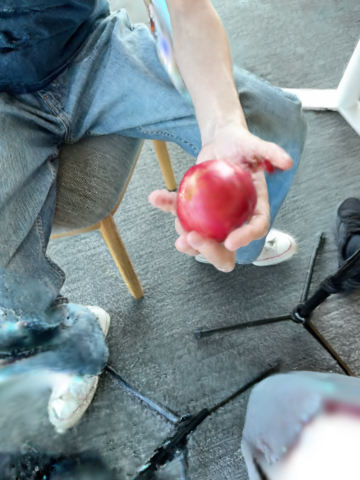} 
        \end{minipage}
        &  
        \begin{minipage}[b]{0.16300\linewidth}
        \includegraphics[width=1\linewidth]{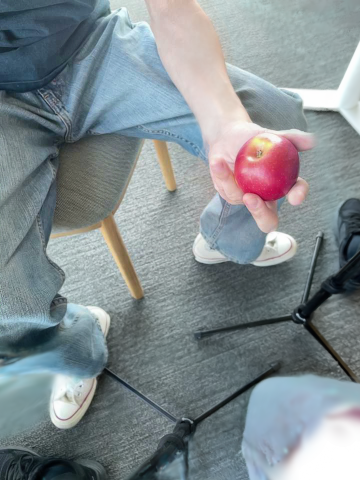} 
        \end{minipage}        
        &  
        \begin{minipage}[b]{0.16300\linewidth}
        \includegraphics[width=1\linewidth]{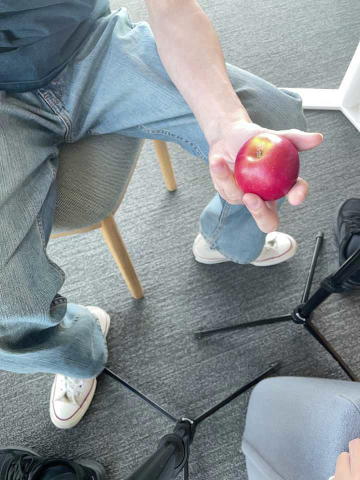}
        \end{minipage}
        \\

        \begin{minipage}[b]{0.16300\linewidth}
        \includegraphics[width=1\linewidth]{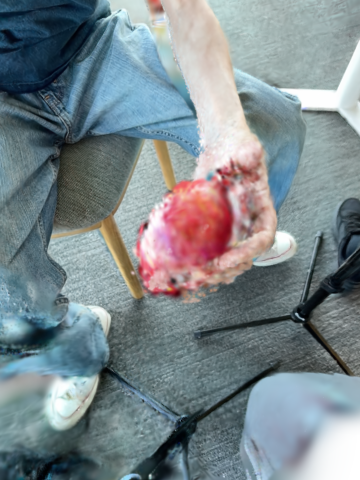} 
        \end{minipage}
        &  
        \begin{minipage}[b]{0.16300\linewidth}
        \includegraphics[width=1\linewidth]{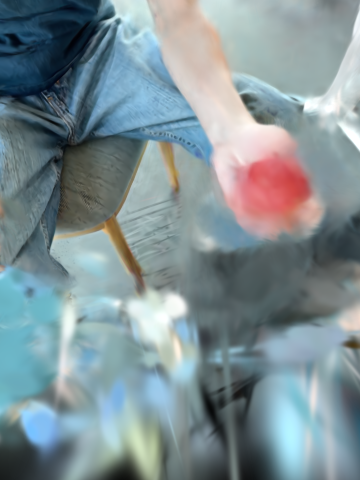} 
        \end{minipage}
        &  

        \begin{minipage}[b]{0.16300\linewidth}
        \includegraphics[width=1\linewidth]{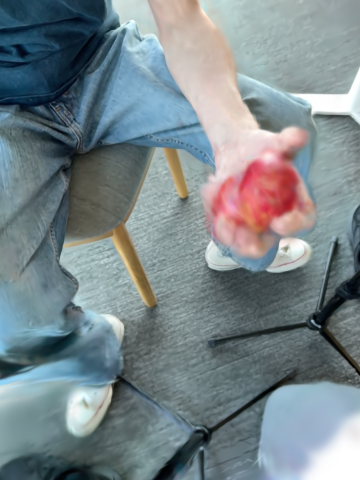} 
        \end{minipage}
        &  
        \begin{minipage}[b]{0.16300\linewidth}
        \includegraphics[width=1\linewidth]{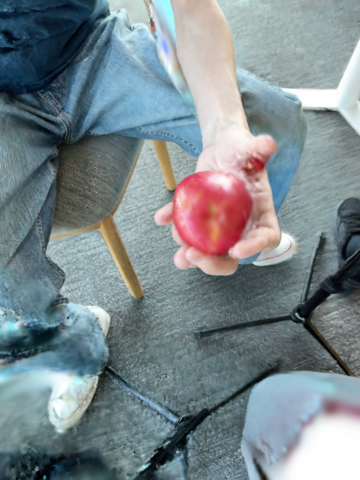} 
        \end{minipage}
        &  
        \begin{minipage}[b]{0.16300\linewidth}
        \includegraphics[width=1\linewidth]{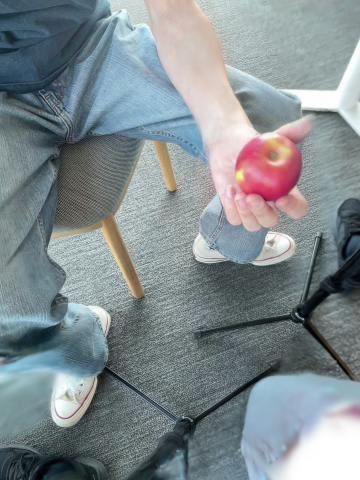}
        \end{minipage}
        &  
        \begin{minipage}[b]{0.16300\linewidth}
        \includegraphics[width=1\linewidth]{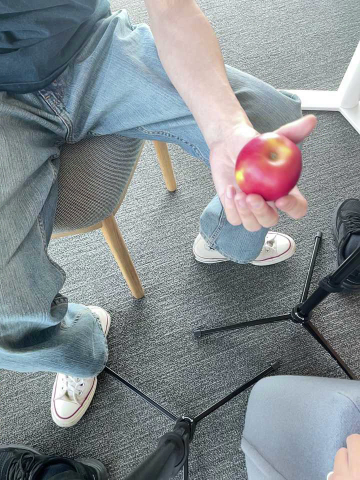}
        \end{minipage}
        \\ 

        \begin{minipage}[b]{0.16300\linewidth}
        \includegraphics[width=1\linewidth]{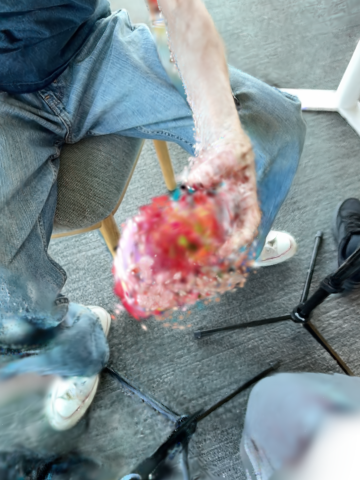} 
        \end{minipage}
        &  
        \begin{minipage}[b]{0.16300\linewidth}
        \includegraphics[width=1\linewidth]{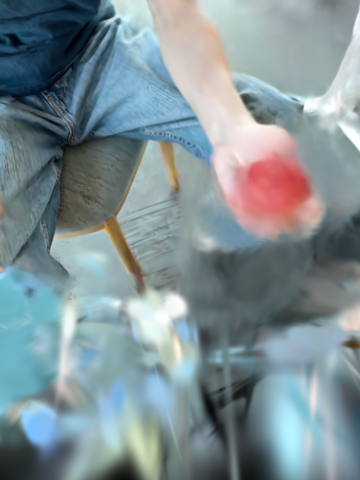} 
        \end{minipage}
        &  

        \begin{minipage}[b]{0.16300\linewidth}
        \includegraphics[width=1\linewidth]{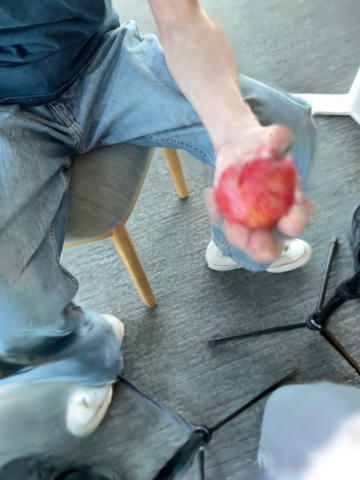} 
        \end{minipage}
        &  
        \begin{minipage}[b]{0.16300\linewidth}
        \includegraphics[width=1\linewidth]{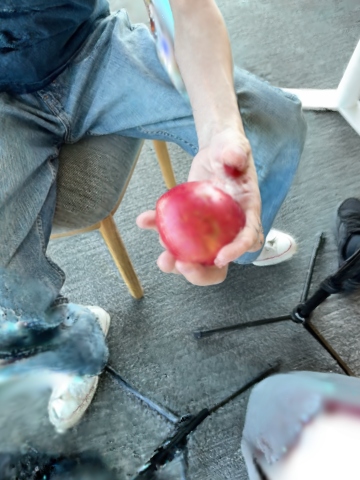} 
        \end{minipage}
        &  
        \begin{minipage}[b]{0.16300\linewidth}
        \includegraphics[width=1\linewidth]{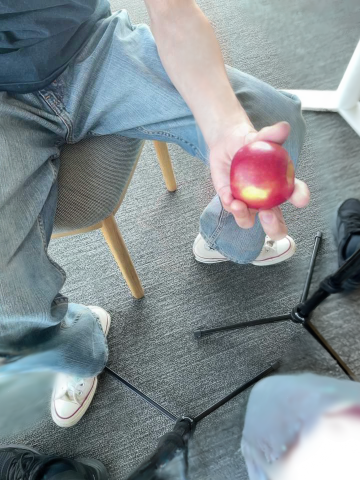}
        \end{minipage}
        &  
        \begin{minipage}[b]{0.16300\linewidth}
        \includegraphics[width=1\linewidth]{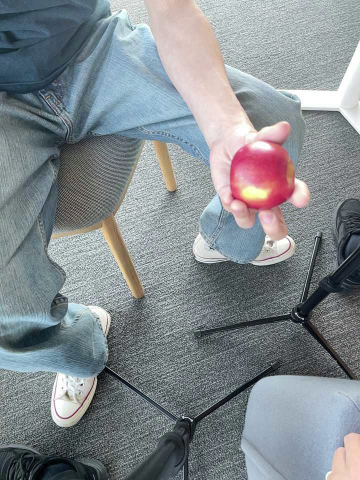}
        \end{minipage}
        \\ 

        \begin{minipage}[b]{0.16300\linewidth}
        \includegraphics[width=1\linewidth]{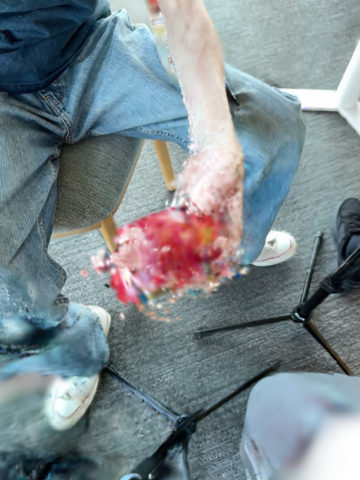} 
        \end{minipage}
        &  
        \begin{minipage}[b]{0.16300\linewidth}
        \includegraphics[width=1\linewidth]{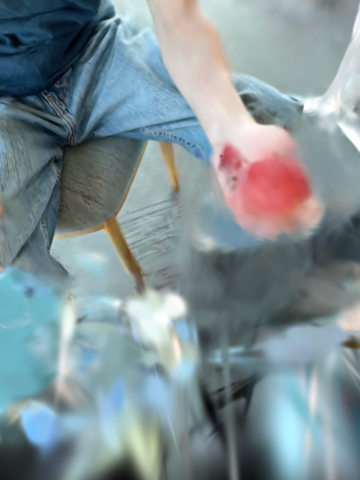} 
        \end{minipage}
        &  

        \begin{minipage}[b]{0.16300\linewidth}
        \includegraphics[width=1\linewidth]{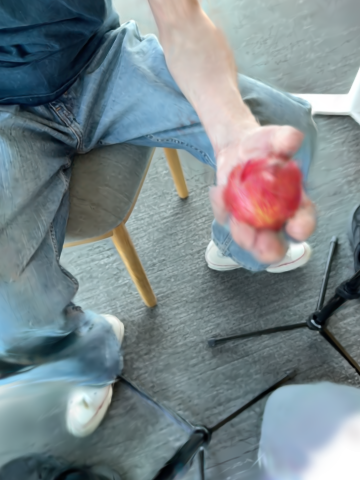} 
        \end{minipage}
        &  
        \begin{minipage}[b]{0.16300\linewidth}
        \includegraphics[width=1\linewidth]{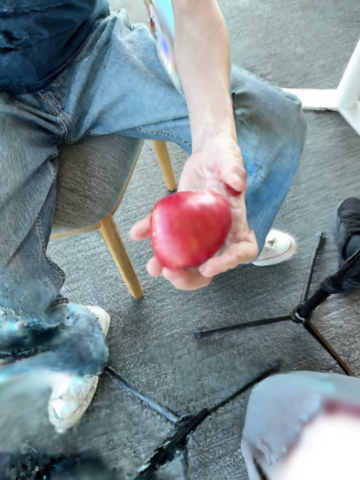} 
        \end{minipage}
        &  
        \begin{minipage}[b]{0.16300\linewidth}
        \includegraphics[width=1\linewidth]{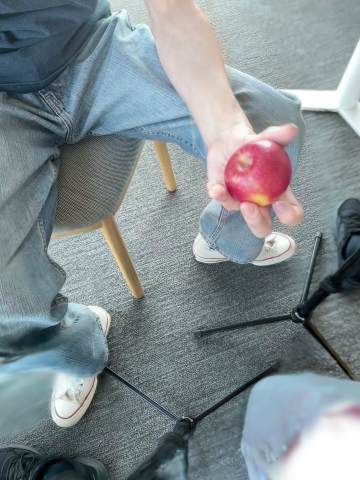}
        \end{minipage}
        &  
        \begin{minipage}[b]{0.16300\linewidth}
        \includegraphics[width=1\linewidth]{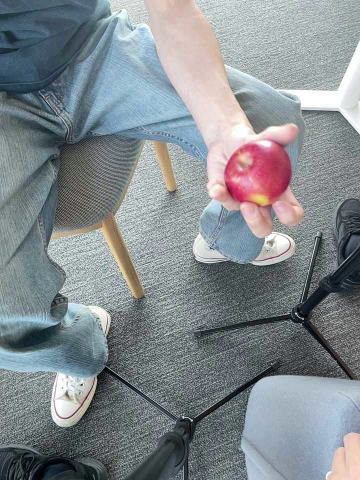}
        \end{minipage}
        \\ 
        
        \begin{minipage}[b]{0.16300\linewidth}
        \includegraphics[width=1\linewidth]{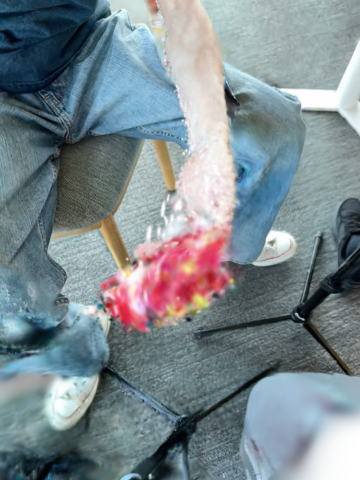} 
        \end{minipage}
        &  
        \begin{minipage}[b]{0.16300\linewidth}
        \includegraphics[width=1\linewidth]{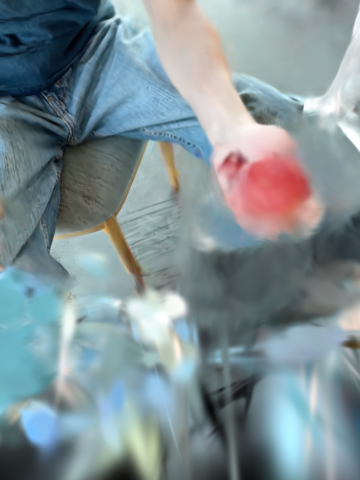} 
        \end{minipage}
        &  

        \begin{minipage}[b]{0.16300\linewidth}
        \includegraphics[width=1\linewidth]{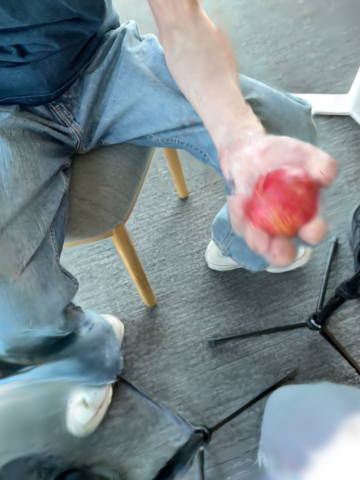} 
        \end{minipage}
        &  
        \begin{minipage}[b]{0.16300\linewidth}
        \includegraphics[width=1\linewidth]{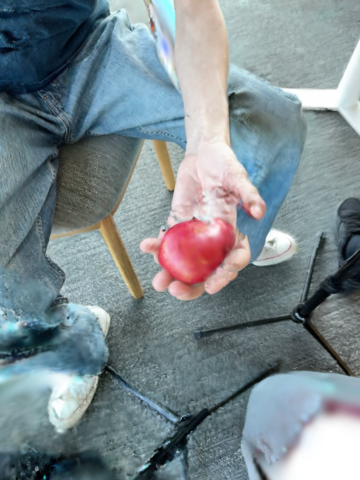} 
        \end{minipage}
        &  
        \begin{minipage}[b]{0.16300\linewidth}
        \includegraphics[width=1\linewidth]{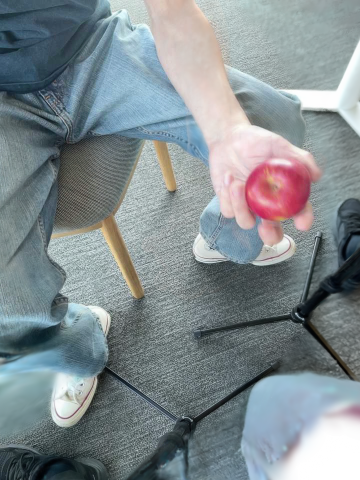}
        \end{minipage}
        &  
        \begin{minipage}[b]{0.16300\linewidth}
        \includegraphics[width=1\linewidth]{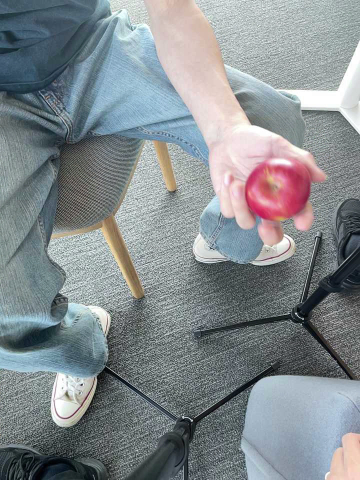}
        \end{minipage}
        \\  
        \makebox[0.16300\linewidth][c]{\footnotesize SoM~\cite{som}} & 
        \makebox[0.16300\linewidth][c]{\footnotesize SplineGS~\cite{splinegs}} & 
        \makebox[0.16300\linewidth][c]{\footnotesize MoSca~\cite{mosca}} & 
        \makebox[0.16300\linewidth][c]{\footnotesize HiMoR~\cite{himor}} & 
        \makebox[0.16300\linewidth][c]{\footnotesize Ours} & 
        \makebox[0.16300\linewidth][c]{\footnotesize GT} \\
    \end{tabular}
    \vspace{-2mm}
    \caption{\textbf{Visual comparison of novel view synthesis on the scene “Apple” of the iPhone dataset \cite{iphone}.} The time interval between adjacent images is ten frames.}
    \label{fig:apple}
\end{figure*}
\begin{figure*}[!h]
    \centering
    \setlength{\tabcolsep}{1.0pt}
    \scriptsize
    \begin{tabular}{cccccc}
        \begin{minipage}[b]{0.16300\linewidth}
        \includegraphics[width=1\linewidth]{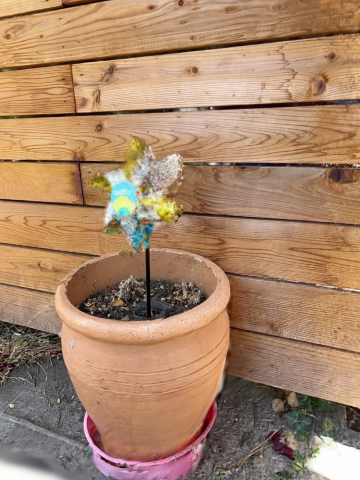}
        \end{minipage}
        &  
        \begin{minipage}[b]{0.16300\linewidth}
        \includegraphics[width=1\linewidth]{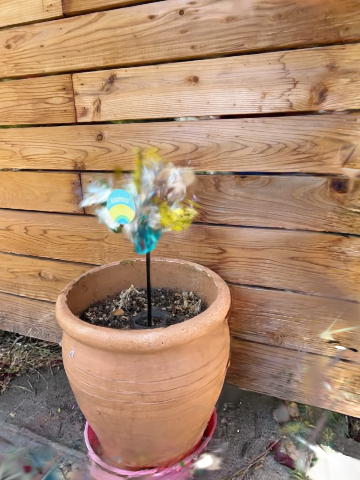} 
        \end{minipage}
        &  
 
        \begin{minipage}[b]{0.16300\linewidth}
        \includegraphics[width=1\linewidth]{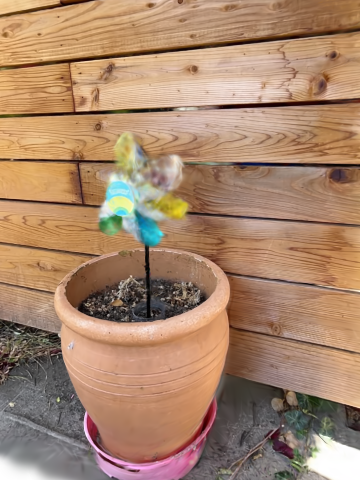} 
        \end{minipage}
        &  
        \begin{minipage}[b]{0.16300\linewidth}
        \includegraphics[width=1\linewidth]{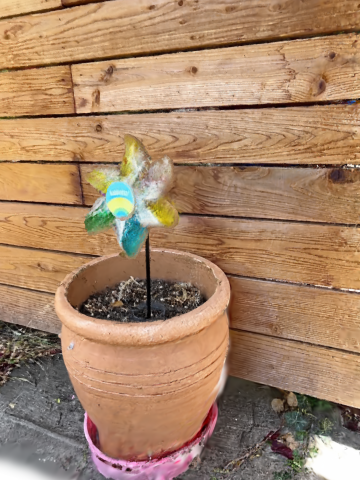} 
        \end{minipage}
        &  
        \begin{minipage}[b]{0.16300\linewidth}
        \includegraphics[width=1\linewidth]{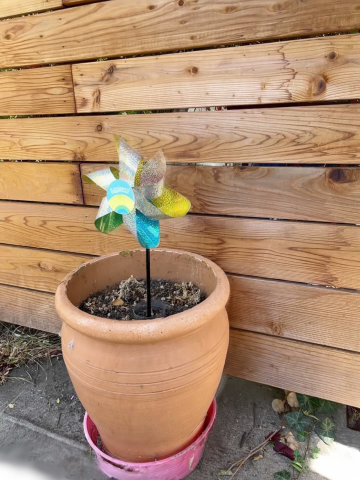} 
        \end{minipage}        
        &  
        \begin{minipage}[b]{0.16300\linewidth}
        \includegraphics[width=1\linewidth]{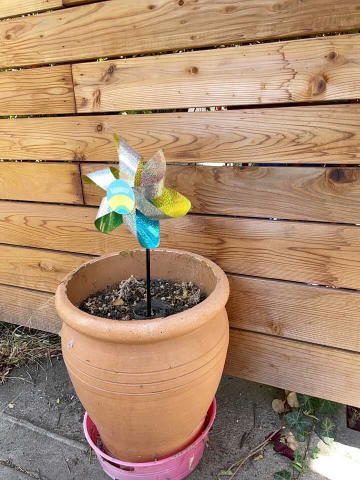}
        \end{minipage}
        \\

        \begin{minipage}[b]{0.16300\linewidth}
        \includegraphics[width=1\linewidth]{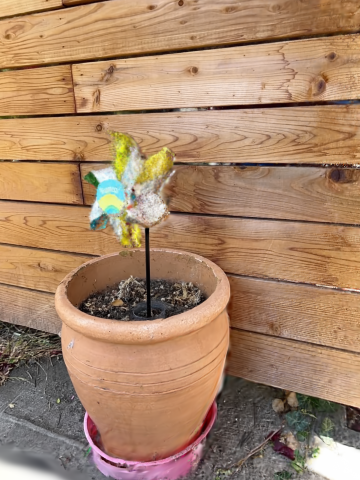} 
        \end{minipage}
        &  
        \begin{minipage}[b]{0.16300\linewidth}
        \includegraphics[width=1\linewidth]{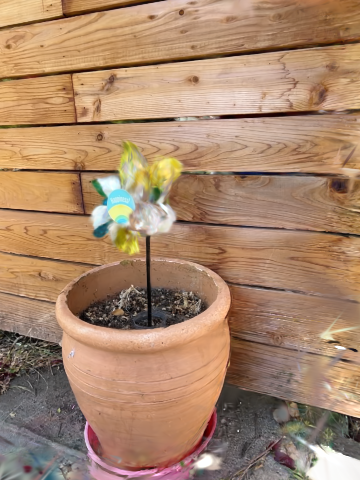} 
        \end{minipage}
        &  

        \begin{minipage}[b]{0.16300\linewidth}
        \includegraphics[width=1\linewidth]{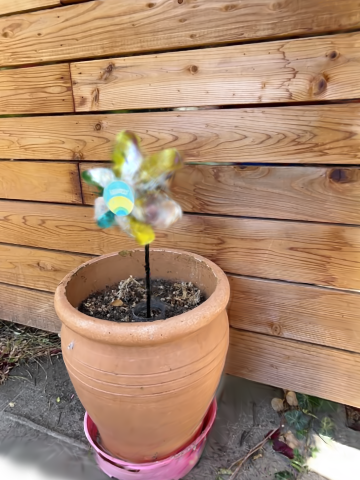} 
        \end{minipage}
        &  
        \begin{minipage}[b]{0.16300\linewidth}
        \includegraphics[width=1\linewidth]{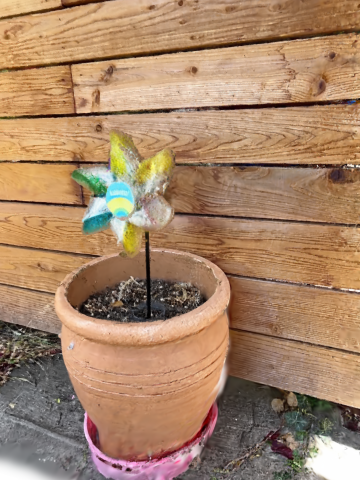} 
        \end{minipage}
        &  
        \begin{minipage}[b]{0.16300\linewidth}
        \includegraphics[width=1\linewidth]{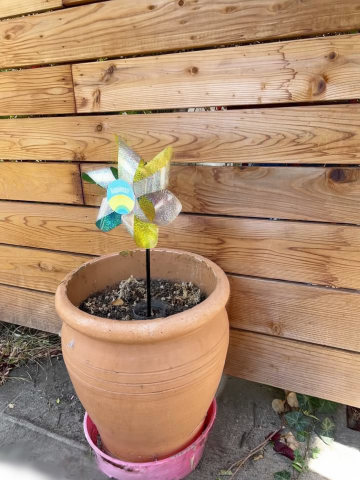}
        \end{minipage}
        &  
        \begin{minipage}[b]{0.16300\linewidth}
        \includegraphics[width=1\linewidth]{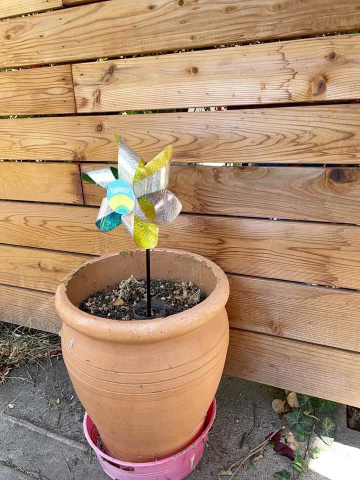}
        \end{minipage}
        \\ 

        \begin{minipage}[b]{0.16300\linewidth}
        \includegraphics[width=1\linewidth]{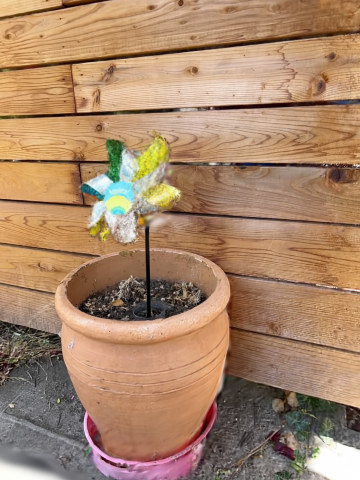} 
        \end{minipage}
        &  
        \begin{minipage}[b]{0.16300\linewidth}
        \includegraphics[width=1\linewidth]{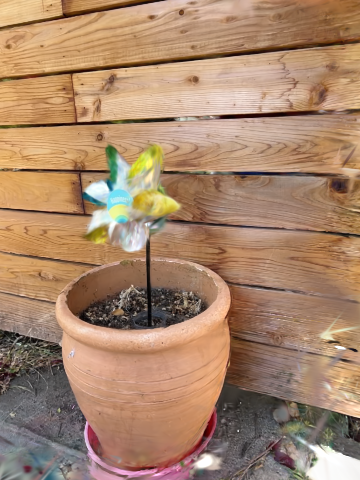} 
        \end{minipage}
        &  

        \begin{minipage}[b]{0.16300\linewidth}
        \includegraphics[width=1\linewidth]{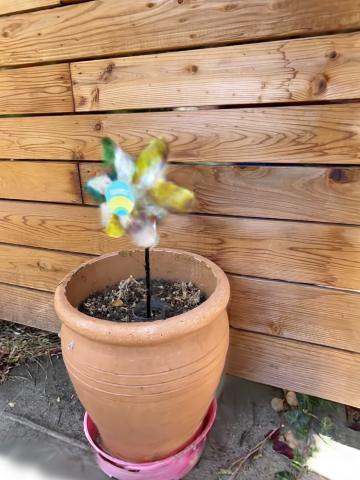} 
        \end{minipage}
        &  
        \begin{minipage}[b]{0.16300\linewidth}
        \includegraphics[width=1\linewidth]{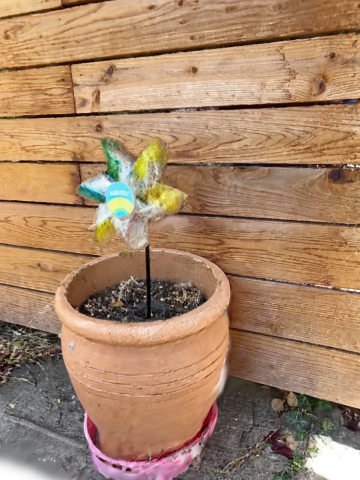} 
        \end{minipage}
        &  
        \begin{minipage}[b]{0.16300\linewidth}
        \includegraphics[width=1\linewidth]{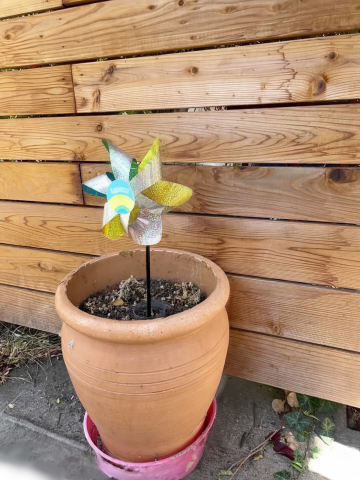}
        \end{minipage}
        &  
        \begin{minipage}[b]{0.16300\linewidth}
        \includegraphics[width=1\linewidth]{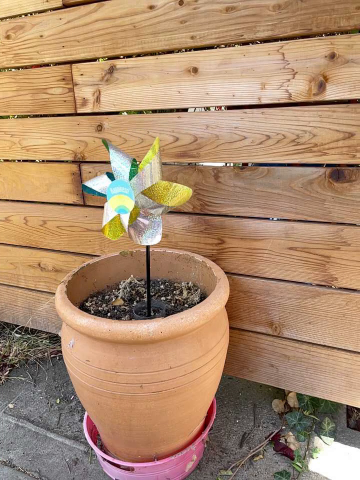}
        \end{minipage}
        \\ 

        \begin{minipage}[b]{0.16300\linewidth}
        \includegraphics[width=1\linewidth]{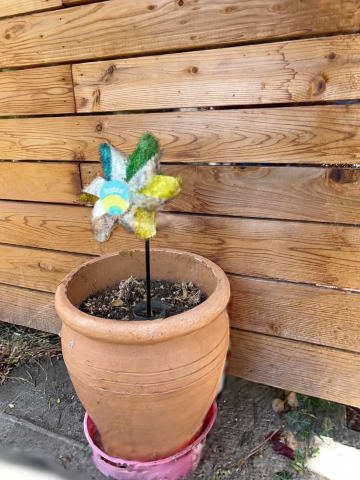} 
        \end{minipage}
        &  
        \begin{minipage}[b]{0.16300\linewidth}
        \includegraphics[width=1\linewidth]{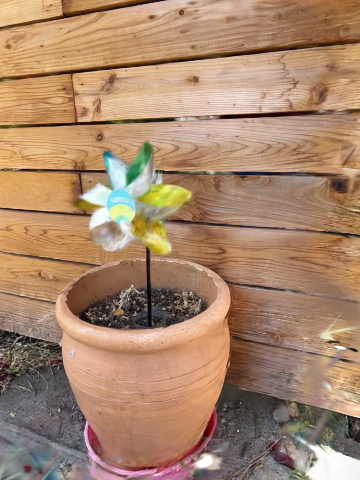} 
        \end{minipage}
        &  

        \begin{minipage}[b]{0.16300\linewidth}
        \includegraphics[width=1\linewidth]{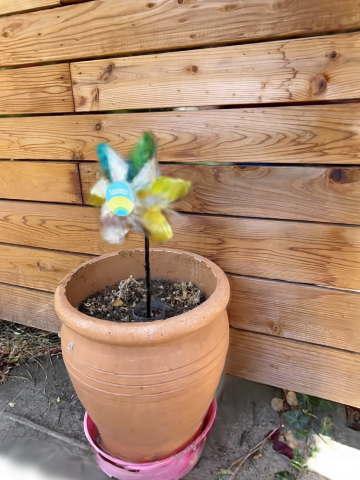} 
        \end{minipage}
        &  
        \begin{minipage}[b]{0.16300\linewidth}
        \includegraphics[width=1\linewidth]{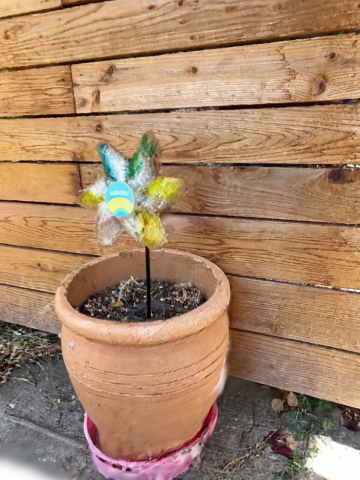} 
        \end{minipage}
        &  
        \begin{minipage}[b]{0.16300\linewidth}
        \includegraphics[width=1\linewidth]{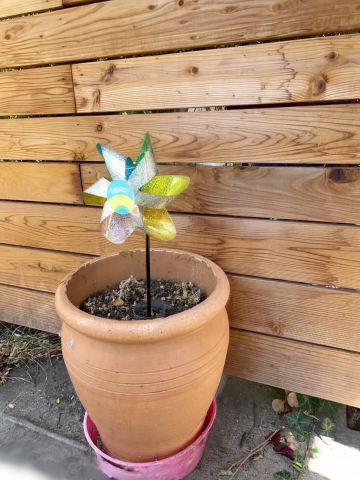}
        \end{minipage}
        &  
        \begin{minipage}[b]{0.16300\linewidth}
        \includegraphics[width=1\linewidth]{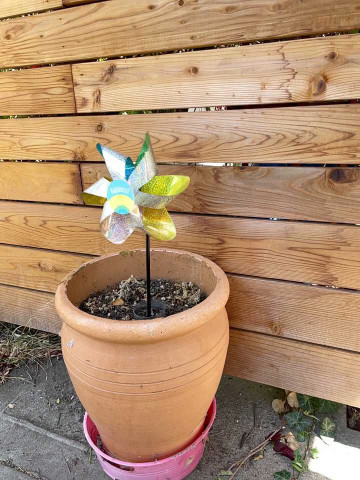}
        \end{minipage}
        \\ 
                \begin{minipage}[b]{0.16300\linewidth}
        \includegraphics[width=1\linewidth]{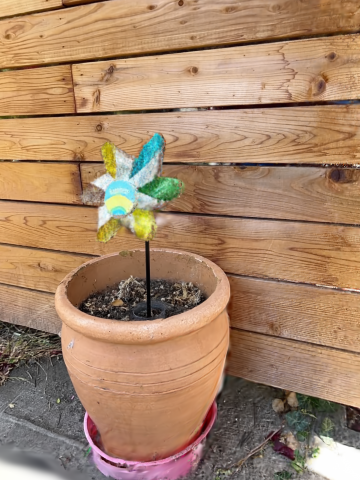} 
        \end{minipage}
        &  
        \begin{minipage}[b]{0.16300\linewidth}
        \includegraphics[width=1\linewidth]{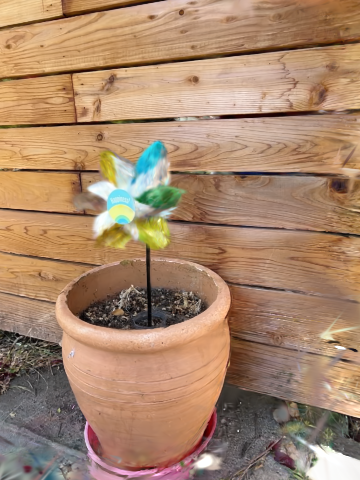} 
        \end{minipage}
        &  

        \begin{minipage}[b]{0.16300\linewidth}
        \includegraphics[width=1\linewidth]{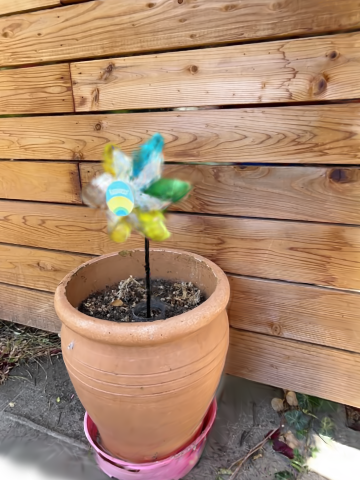} 
        \end{minipage}
        &  
        \begin{minipage}[b]{0.16300\linewidth}
        \includegraphics[width=1\linewidth]{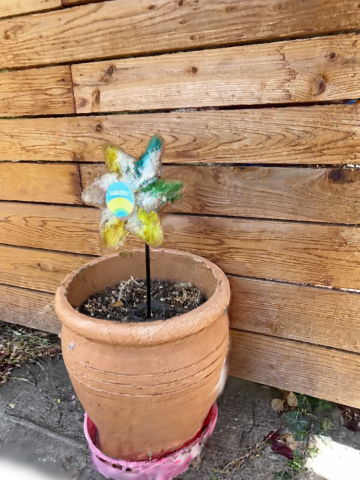} 
        \end{minipage}
        &  
        \begin{minipage}[b]{0.16300\linewidth}
        \includegraphics[width=1\linewidth]{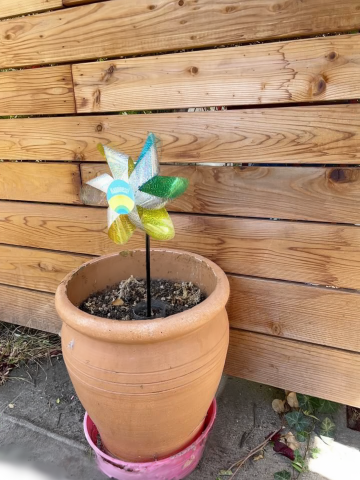}
        \end{minipage}
        &  
        \begin{minipage}[b]{0.16300\linewidth}
        \includegraphics[width=1\linewidth]{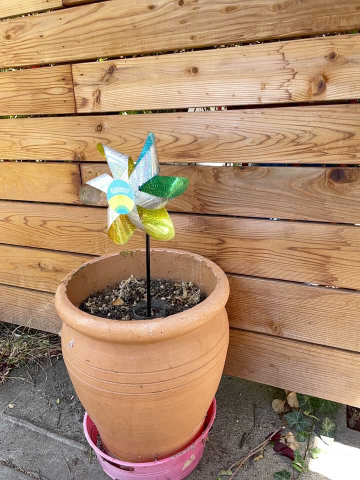}
        \end{minipage}
        \\  
        \makebox[0.16300\linewidth][c]{\footnotesize SoM~\cite{som}} & 
        \makebox[0.16300\linewidth][c]{\footnotesize SplineGS~\cite{splinegs}} & 
        \makebox[0.16300\linewidth][c]{\footnotesize MoSca~\cite{mosca}} & 
        \makebox[0.16300\linewidth][c]{\footnotesize HiMoR~\cite{himor}} & 
        \makebox[0.16300\linewidth][c]{\footnotesize Ours} & 
        \makebox[0.16300\linewidth][c]{\footnotesize GT} \\
    \end{tabular}
    \vspace{-2mm}
    \caption{\textbf{Visual comparison of novel view synthesis on the scene “Paper-windmill” of the iPhone dataset \cite{iphone}.} The time interval between adjacent images is ten frames.}
    \label{fig:paperwindmill}
\end{figure*}
\begin{figure*}[!t]
    \centering
    \setlength{\tabcolsep}{1.0pt}
    \scriptsize
    \begin{tabular}{cccccc}
        \begin{minipage}[b]{0.16260\linewidth}
        \includegraphics[width=1\linewidth]{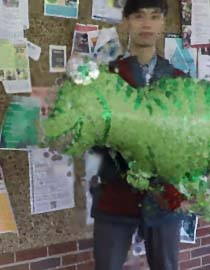}
        \end{minipage}
        &  
        \begin{minipage}[b]{0.16260\linewidth}
        \includegraphics[width=1\linewidth]{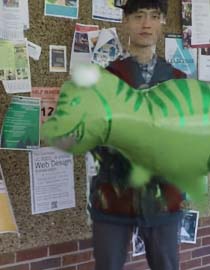} 
        \end{minipage}
        &  

        \begin{minipage}[b]{0.16260\linewidth}
        \includegraphics[width=1\linewidth]{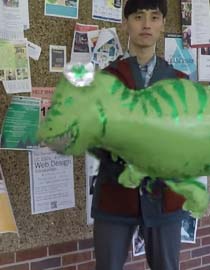} 
        \end{minipage}
        &  
        \begin{minipage}[b]{0.16260\linewidth}
        \includegraphics[width=1\linewidth]{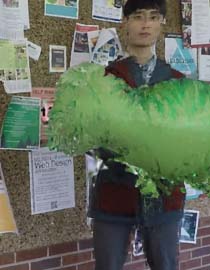} 
        \end{minipage}
        &  
        \begin{minipage}[b]{0.16260\linewidth}
        \includegraphics[width=1\linewidth]{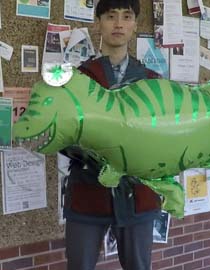} 
        \end{minipage}        
        &  
        \begin{minipage}[b]{0.16260\linewidth}
        \includegraphics[width=1\linewidth]{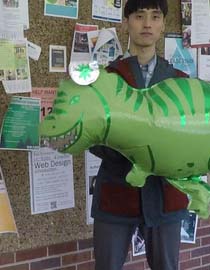}
        \end{minipage}
        \\
        \begin{minipage}[b]{0.16260\linewidth}
        \includegraphics[width=1\linewidth]{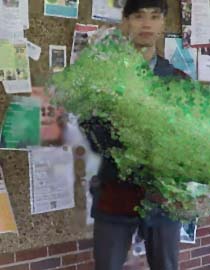}
        \end{minipage}
        &  
        \begin{minipage}[b]{0.16260\linewidth}
        \includegraphics[width=1\linewidth]{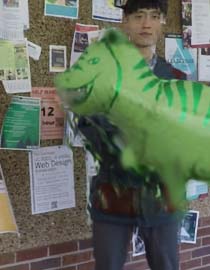} 
        \end{minipage}
        &  

        \begin{minipage}[b]{0.16260\linewidth}
        \includegraphics[width=1\linewidth]{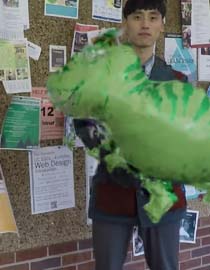} 
        \end{minipage}
        &  
        \begin{minipage}[b]{0.16260\linewidth}
        \includegraphics[width=1\linewidth]{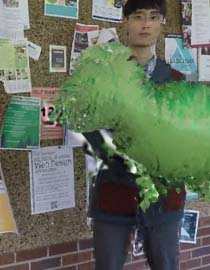} 
        \end{minipage}
        &  
        \begin{minipage}[b]{0.16260\linewidth}
        \includegraphics[width=1\linewidth]{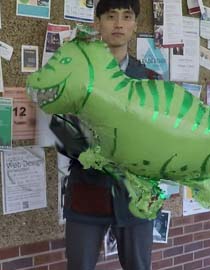} 
        \end{minipage}        
        &  
        \begin{minipage}[b]{0.16260\linewidth}
        \includegraphics[width=1\linewidth]{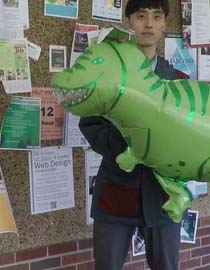}
        \end{minipage}
        \\
        \begin{minipage}[b]{0.16260\linewidth}
        \includegraphics[width=1\linewidth]{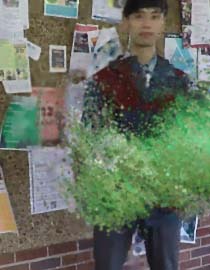}
        \end{minipage}
        &  
        \begin{minipage}[b]{0.16260\linewidth}
        \includegraphics[width=1\linewidth]{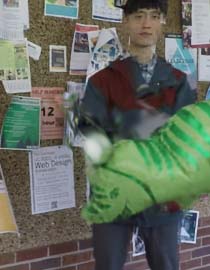} 
        \end{minipage}
        &  

        \begin{minipage}[b]{0.16260\linewidth}
        \includegraphics[width=1\linewidth]{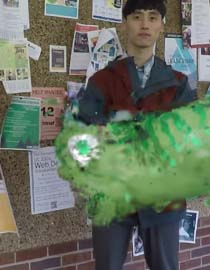} 
        \end{minipage}
        &  
        \begin{minipage}[b]{0.16260\linewidth}
        \includegraphics[width=1\linewidth]{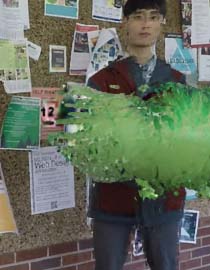} 
        \end{minipage}
        &  
        \begin{minipage}[b]{0.16260\linewidth}
        \includegraphics[width=1\linewidth]{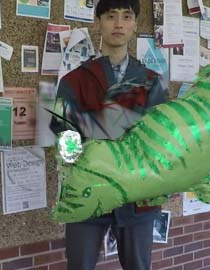} 
        \end{minipage}        
        &  
        \begin{minipage}[b]{0.16260\linewidth}
        \includegraphics[width=1\linewidth]{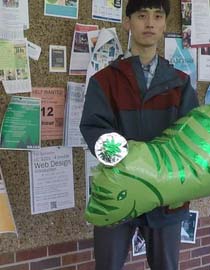}
        \end{minipage}
        \\
        \begin{minipage}[b]{0.16260\linewidth}
        \includegraphics[width=1\linewidth]{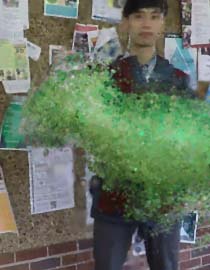}
        \end{minipage}
        &  
        \begin{minipage}[b]{0.16260\linewidth}
        \includegraphics[width=1\linewidth]{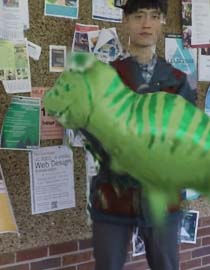} 
        \end{minipage}
        &  

        \begin{minipage}[b]{0.16260\linewidth}
        \includegraphics[width=1\linewidth]{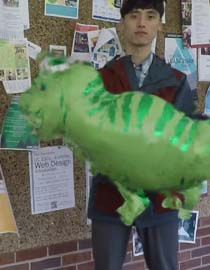} 
        \end{minipage}
        &  
        \begin{minipage}[b]{0.16260\linewidth}
        \includegraphics[width=1\linewidth]{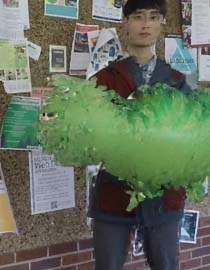} 
        \end{minipage}
        &  
        \begin{minipage}[b]{0.16260\linewidth}
        \includegraphics[width=1\linewidth]{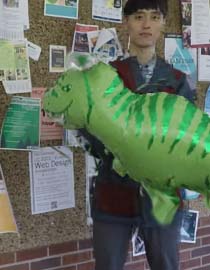} 
        \end{minipage}        
        &  
        \begin{minipage}[b]{0.16260\linewidth}
        \includegraphics[width=1\linewidth]{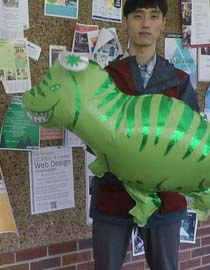}
        \end{minipage}
        \\
        \begin{minipage}[b]{0.16260\linewidth}
        \includegraphics[width=1\linewidth]{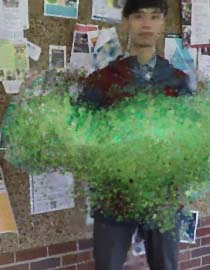}
        \end{minipage}
        &  
        \begin{minipage}[b]{0.16260\linewidth}
        \includegraphics[width=1\linewidth]{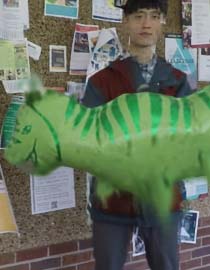} 
        \end{minipage}
        &  

        \begin{minipage}[b]{0.16260\linewidth}
        \includegraphics[width=1\linewidth]{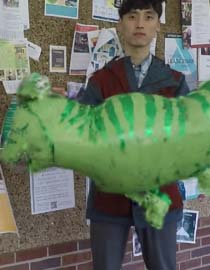} 
        \end{minipage}
        &  
        \begin{minipage}[b]{0.16260\linewidth}
        \includegraphics[width=1\linewidth]{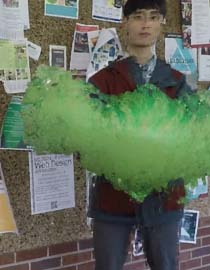} 
        \end{minipage}
        &  
        \begin{minipage}[b]{0.16260\linewidth}
        \includegraphics[width=1\linewidth]{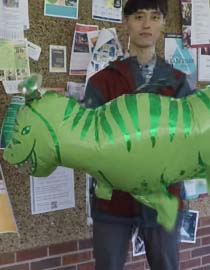} 
        \end{minipage}        
        &  
        \begin{minipage}[b]{0.16260\linewidth}
        \includegraphics[width=1\linewidth]{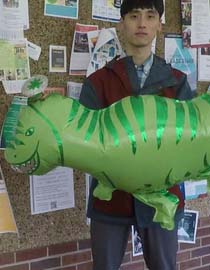}
        \end{minipage}
        \\
        \begin{minipage}[b]{0.16260\linewidth}
        \includegraphics[width=1\linewidth]{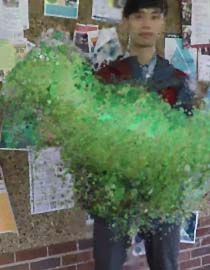}
        \end{minipage}
        &  
        \begin{minipage}[b]{0.16260\linewidth}
        \includegraphics[width=1\linewidth]{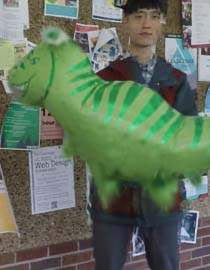} 
        \end{minipage}
        &  

        \begin{minipage}[b]{0.16260\linewidth}
        \includegraphics[width=1\linewidth]{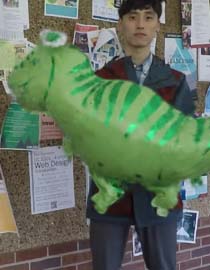} 
        \end{minipage}
        &  
        \begin{minipage}[b]{0.16260\linewidth}
        \includegraphics[width=1\linewidth]{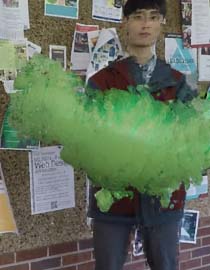} 
        \end{minipage}
        &  
        \begin{minipage}[b]{0.16260\linewidth}
        \includegraphics[width=1\linewidth]{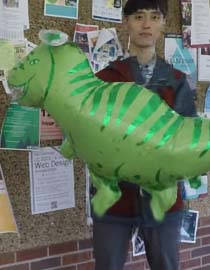} 
        \end{minipage}        
        &  
        \begin{minipage}[b]{0.16260\linewidth}
        \includegraphics[width=1\linewidth]{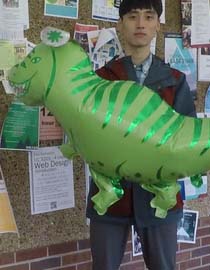}
        \end{minipage}
        \\
        \makebox[0.16260\linewidth][c]{\footnotesize SoM~\cite{som} } & 
        \makebox[0.16260\linewidth][c]{\footnotesize SplineGS~\cite{splinegs}} & 

        \makebox[0.16260\linewidth][c]{\footnotesize MoSca~\cite{mosca}} & 
        \makebox[0.16260\linewidth][c]{\footnotesize HiMoR~\cite{himor}} & 
        \makebox[0.16260\linewidth][c]{\footnotesize Ours} & 
        \makebox[0.16260\linewidth][c]{\footnotesize GT} \\
    \end{tabular}
    \vspace{-2mm}
    \caption{\textbf{Visual comparison of novel view synthesis on the scene “Balloon1” of the NVIDIA dataset \cite{nvidia}.} The time interval between adjacent images is two frames.}
    \label{fig:balloon1}
\end{figure*}


\begin{figure*}[!t]
    \centering
    \setlength{\tabcolsep}{1.0pt}
    \scriptsize
    \begin{tabular}{cccccc}
        \begin{minipage}[b]{0.16260\linewidth}
        \includegraphics[width=1\linewidth]{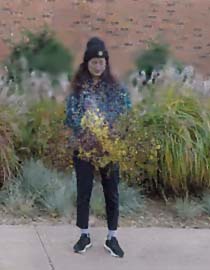}
        \end{minipage}
        &  
        \begin{minipage}[b]{0.16260\linewidth}
        \includegraphics[width=1\linewidth]{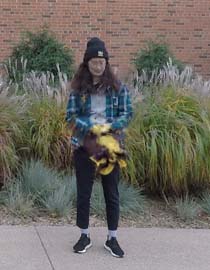} 
        \end{minipage}
        &  

        \begin{minipage}[b]{0.16260\linewidth}
        \includegraphics[width=1\linewidth]{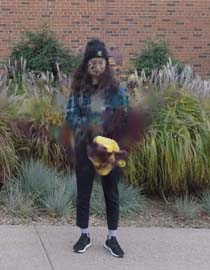} 
        \end{minipage}
        &  
        \begin{minipage}[b]{0.16260\linewidth}
        \includegraphics[width=1\linewidth]{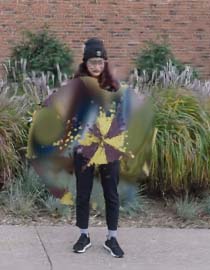} 
        \end{minipage}
        &  
        \begin{minipage}[b]{0.16260\linewidth}
        \includegraphics[width=1\linewidth]{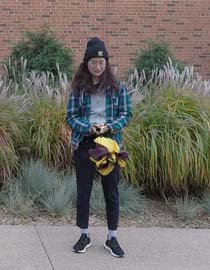} 
        \end{minipage}        
        &  
        \begin{minipage}[b]{0.16260\linewidth}
        \includegraphics[width=1\linewidth]{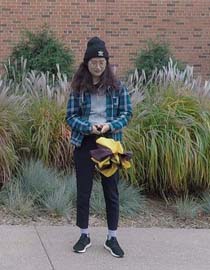}
        \end{minipage}
        \\
        \begin{minipage}[b]{0.16260\linewidth}
        \includegraphics[width=1\linewidth]{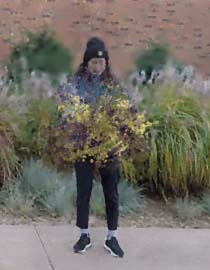}
        \end{minipage}
        &  
        \begin{minipage}[b]{0.16260\linewidth}
        \includegraphics[width=1\linewidth]{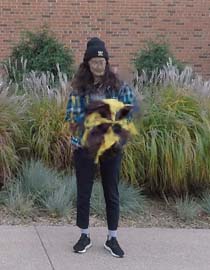} 
        \end{minipage}
        &  

        \begin{minipage}[b]{0.16260\linewidth}
        \includegraphics[width=1\linewidth]{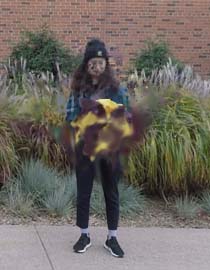} 
        \end{minipage}
        &  
        \begin{minipage}[b]{0.16260\linewidth}
        \includegraphics[width=1\linewidth]{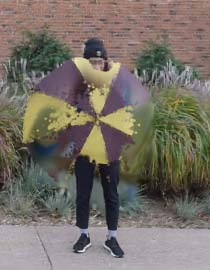} 
        \end{minipage}
        &  
        \begin{minipage}[b]{0.16260\linewidth}
        \includegraphics[width=1\linewidth]{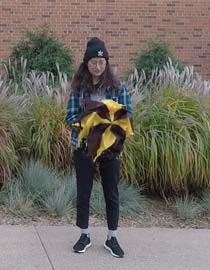} 
        \end{minipage}        
        &  
        \begin{minipage}[b]{0.16260\linewidth}
        \includegraphics[width=1\linewidth]{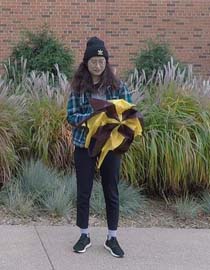}
        \end{minipage}
        \\
        \begin{minipage}[b]{0.16260\linewidth}
        \includegraphics[width=1\linewidth]{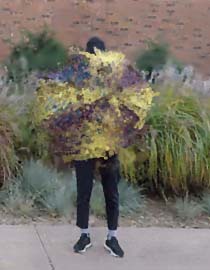}
        \end{minipage}
        &  
        \begin{minipage}[b]{0.16260\linewidth}
        \includegraphics[width=1\linewidth]{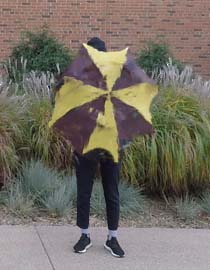} 
        \end{minipage}
        &  

        \begin{minipage}[b]{0.16260\linewidth}
        \includegraphics[width=1\linewidth]{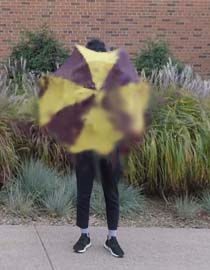} 
        \end{minipage}
        &  
        \begin{minipage}[b]{0.16260\linewidth}
        \includegraphics[width=1\linewidth]{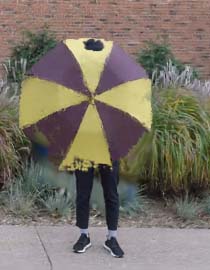} 
        \end{minipage}
        &  
        \begin{minipage}[b]{0.16260\linewidth}
        \includegraphics[width=1\linewidth]{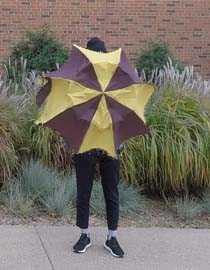} 
        \end{minipage}        
        &  
        \begin{minipage}[b]{0.16260\linewidth}
        \includegraphics[width=1\linewidth]{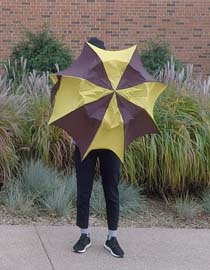}
        \end{minipage}
        \\
        \begin{minipage}[b]{0.16260\linewidth}
        \includegraphics[width=1\linewidth]{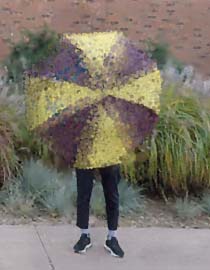}
        \end{minipage}
        &  
        \begin{minipage}[b]{0.16260\linewidth}
        \includegraphics[width=1\linewidth]{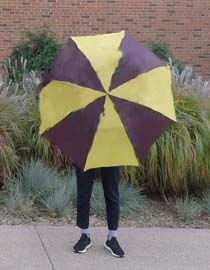} 
        \end{minipage}
        &  

        \begin{minipage}[b]{0.16260\linewidth}
        \includegraphics[width=1\linewidth]{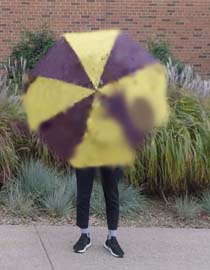} 
        \end{minipage}
        &  
        \begin{minipage}[b]{0.16260\linewidth}
        \includegraphics[width=1\linewidth]{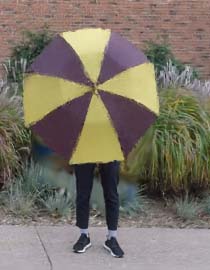} 
        \end{minipage}
        &  
        \begin{minipage}[b]{0.16260\linewidth}
        \includegraphics[width=1\linewidth]{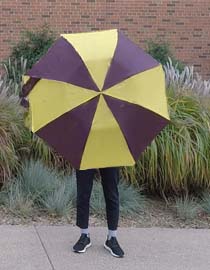} 
        \end{minipage}        
        &  
        \begin{minipage}[b]{0.16260\linewidth}
        \includegraphics[width=1\linewidth]{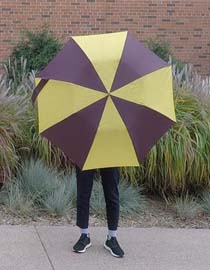}
        \end{minipage}
        \\
        \begin{minipage}[b]{0.16260\linewidth}
        \includegraphics[width=1\linewidth]{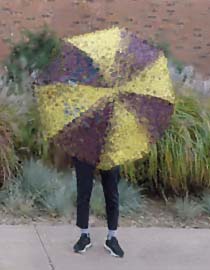}
        \end{minipage}
        &  
        \begin{minipage}[b]{0.16260\linewidth}
        \includegraphics[width=1\linewidth]{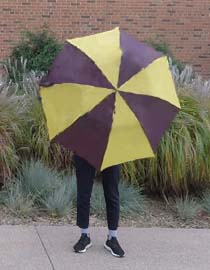} 
        \end{minipage}
        &  

        \begin{minipage}[b]{0.16260\linewidth}
        \includegraphics[width=1\linewidth]{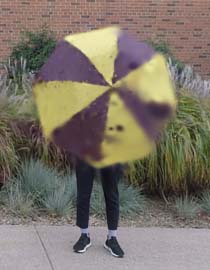} 
        \end{minipage}
        &  
        \begin{minipage}[b]{0.16260\linewidth}
        \includegraphics[width=1\linewidth]{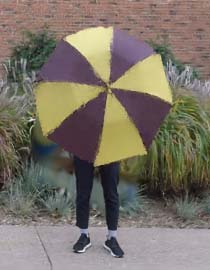} 
        \end{minipage}
        &  
        \begin{minipage}[b]{0.16260\linewidth}
        \includegraphics[width=1\linewidth]{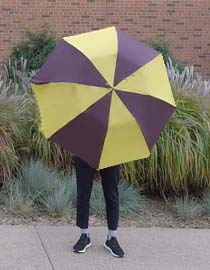} 
        \end{minipage}        
        &  
        \begin{minipage}[b]{0.16260\linewidth}
        \includegraphics[width=1\linewidth]{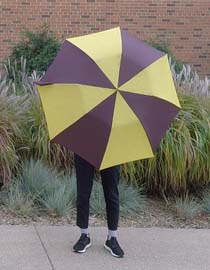}
        \end{minipage}
        \\
        \begin{minipage}[b]{0.16260\linewidth}
        \includegraphics[width=1\linewidth]{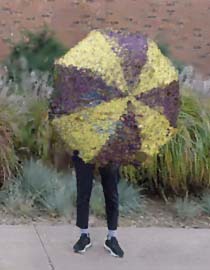}
        \end{minipage}
        &  
        \begin{minipage}[b]{0.16260\linewidth}
        \includegraphics[width=1\linewidth]{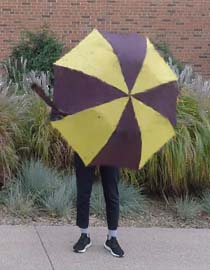} 
        \end{minipage}
        &  

        \begin{minipage}[b]{0.16260\linewidth}
        \includegraphics[width=1\linewidth]{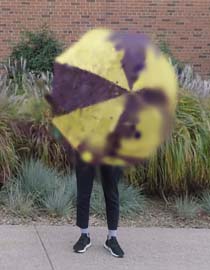} 
        \end{minipage}
        &  
        \begin{minipage}[b]{0.16260\linewidth}
        \includegraphics[width=1\linewidth]{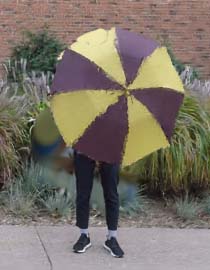} 
        \end{minipage}
        &  
        \begin{minipage}[b]{0.16260\linewidth}
        \includegraphics[width=1\linewidth]{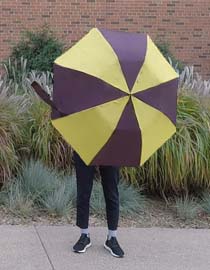} 
        \end{minipage}        
        &  
        \begin{minipage}[b]{0.16260\linewidth}
        \includegraphics[width=1\linewidth]{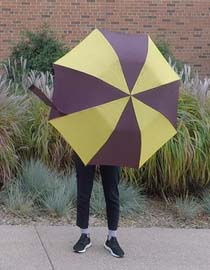}
        \end{minipage}
        \\
        \makebox[0.16260\linewidth][c]{\footnotesize SoM~\cite{som} } & 
        \makebox[0.16260\linewidth][c]{\footnotesize SplineGS~\cite{splinegs}} & 

        \makebox[0.16260\linewidth][c]{\footnotesize MoSca~\cite{mosca}} & 
        \makebox[0.16260\linewidth][c]{\footnotesize HiMoR~\cite{himor}} & 
        \makebox[0.16260\linewidth][c]{\footnotesize Ours} & 
        \makebox[0.16260\linewidth][c]{\footnotesize GT} \\
    \end{tabular}
    \vspace{-2mm}
    \caption{\textbf{Visual comparison of novel view synthesis on the scene “Umbrella” of the NVIDIA dataset \cite{nvidia}.} The time interval between adjacent images is two frames.}
    \label{fig:umbrella}
\end{figure*}

\end{document}